\documentclass[10pt,twocolumn,letterpaper]{article}

\usepackage{iccv}              

\usepackage{indentfirst}

\definecolor{iccvblue}{rgb}{0.21,0.49,0.74}

\usepackage{amsmath}
\usepackage{amssymb}
\usepackage{mathtools}
\usepackage{amsthm}
\usepackage{multirow}
\usepackage{colortbl}
\usepackage{adjustbox}

\definecolor{maroonx}{RGB}{195,18,48}
\usepackage[colorlinks=True,
            linkcolor=maroonx,
            anchorcolor=blue,
            citecolor=iccvblue,
            ]{hyperref}
\usepackage{tcolorbox}
\newcommand{\cc}{\cellcolor{gray!20}}
\definecolor{darkred}{RGB}{192, 0, 0}
\definecolor{darkgreen}{RGB}{103, 174, 64}
\definecolor{darkblue}{RGB}{56, 84, 146}

\usepackage{algorithm}
\usepackage{algpseudocode}
\usepackage{amsmath, amssymb}

\def\paperID{15001} 
\def\confName{ICCV}
\def\confYear{2025}

\newcommand{\RebuttalRevision}[1]{\textcolor{black}{#1}}

\title{ONLY: One-Layer Intervention Sufficiently Mitigates Hallucinations in \\Large Vision-Language Models}

\author{Zifu Wan\thanks{Equal contribution. \, Contact: {\small \texttt{zifuw@cs.cmu.edu}}.}\hspace{0.45em}\quad Ce Zhang\footnotemark[1]\hspace{0.45em}\quad Silong Yong \quad Martin Q. Ma\quad Simon Stepputtis\\ \vspace{1mm} Louis-Philippe Morency\quad 
Deva Ramanan\quad Katia Sycara\quad Yaqi Xie \\
Carnegie Mellon University \vspace{-3mm}}

\begin{document}
\maketitle
\begin{abstract}
\looseness=-1
Recent Large Vision-Language Models (LVLMs) have introduced a new paradigm for understanding and reasoning about image input through textual responses. Although they have achieved remarkable performance across a range of multi-modal tasks, they face the persistent challenge of \textit{hallucination}, which introduces practical weaknesses and raises concerns about their reliable deployment in real-world applications. Existing work has explored contrastive decoding approaches to mitigate this issue, where the output of the original LVLM is compared and contrasted with that of a perturbed version. However, these methods require two or more queries that slow down LVLM response generation, making them less suitable for real-time applications. To overcome this limitation, we propose ONLY, a training-free decoding approach that requires only a single query and a one-layer intervention during decoding, enabling efficient real-time deployment. Specifically, we enhance textual outputs by selectively amplifying crucial textual information using a text-to-visual entropy ratio for each token.
Extensive experimental results demonstrate that our proposed ONLY consistently outperforms state-of-the-art methods across various benchmarks while requiring minimal implementation effort and computational cost. Code is available at \url{https://github.com/zifuwan/ONLY}.
\end{abstract}

\vspace{-7pt}
\section{Introduction}
\label{sec:intro}
\begin{figure}[t]
  \centering
  \includegraphics[width=0.9\linewidth]{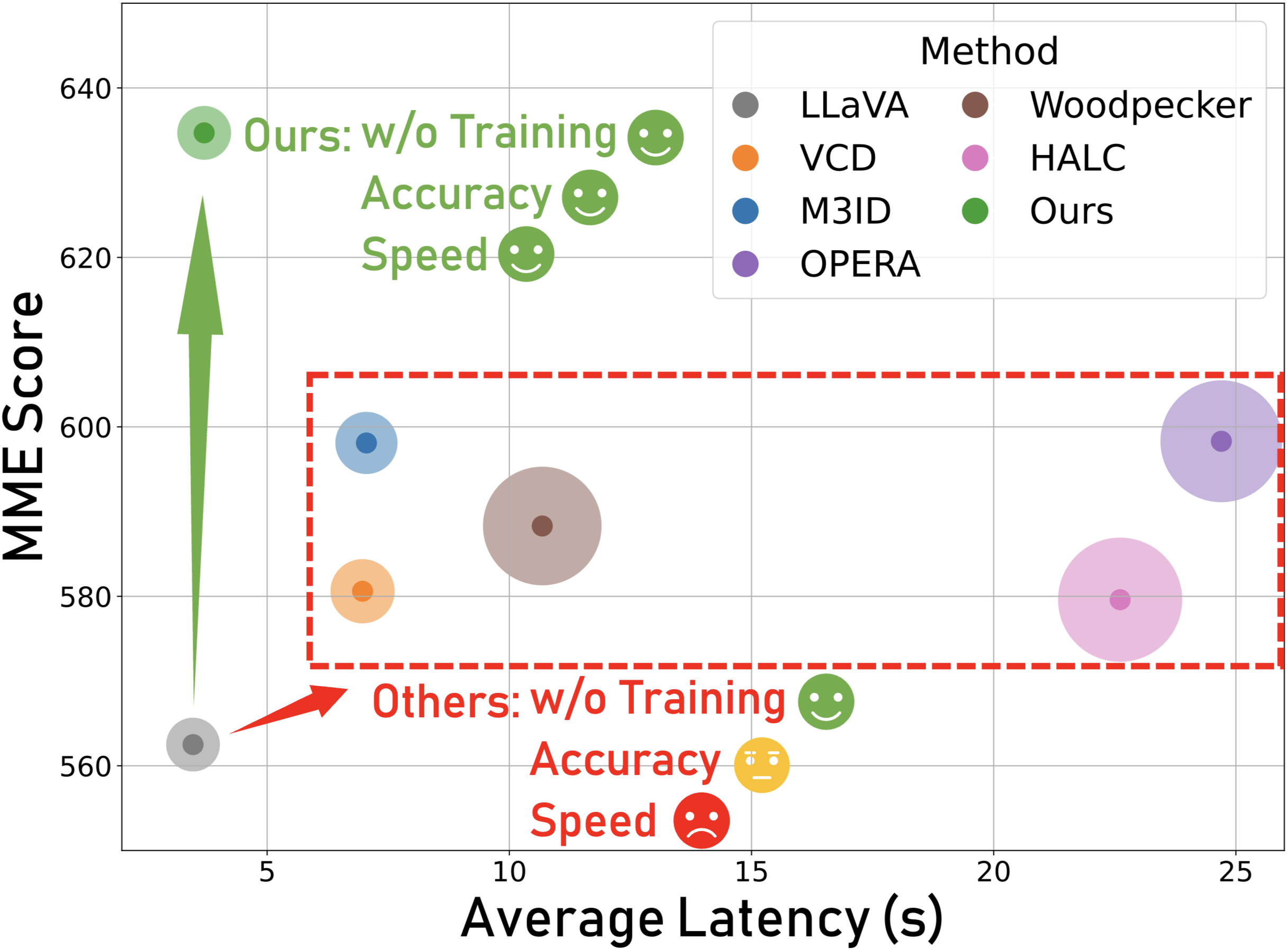}
  \vspace{-5pt}
  \caption{\textbf{Comparisons of accuracy and inference speed of multiple hallucination mitigation approaches.} The size of bubbles stands for the GPU memory consumption. Our method effectively mitigates hallucination with only 0.07$\times$ extra time.}
  \label{fig:teaser}
  \vspace{-10pt}
\end{figure}


\looseness=-1
Recent advances in large vision-language models (LVLMs), which expand the capabilities of large language models (LLMs) to visual understanding and reasoning~\citep{li2023blip,bai2023qwen,ye2024mplug}, have demonstrated exceptional performance across various vision-language tasks, such as object detection~\citep{wu2024dettoolchain,zang2024contextual}, segmentation~\cite{wan2025instructpart,lai2024lisa}, and image captioning~\citep{lin2014microsoft,rohrbach2018object}. However, a persistent challenge with current LVLMs is their tendency to generate hallucinated content, where the generated responses do not align accurately with the actual image input~\citep{liu2024survey}. 
This can significantly impact the reliability of LVLMs in real-world applications where precise visual interpretation is essential~\citep{liu2024survey,bai2024hallucination,chen2024detecting}. Therefore, addressing hallucinations in LVLMs is crucial to ensuring their safe and effective deployment in critical domains.

\looseness=-1
To alleviate the hallucination problem, early work identified biased training sets as a critical cause and, as a result, attempted to establish curated training datasets and adopt robust fine-tuning techniques~\citep{zhang2024reflective,chen2023mitigating,sun2023aligning}. However, their reliance on additional data and the need for fine-tuning large-scale models make these approaches time-consuming and impractical for individual users. Another common approach is contrastive decoding~\citep{li2023contrastive}, which eliminates the need for costly training by directly intervening in the decoding process during inference. Specifically, these methods typically introduce a distorted set of inputs, 
and contrast their respective token predictions with the predictions from original data to mitigate undesired hallucinations~\citep{chuang2024dola,wang2024mitigating2,favero2024multi,leng2024mitigating}. Although existing contrastive decoding-based approaches achieve notable performance improvements, they require multiple LVLM queries to process both the original and distorted inputs, resulting in response times that are twice as long, or more, making them less suitable for real-time applications~\citep{favero2024multi,leng2024mitigating,chen2024image}.

To illustrate this, we analyze the performance-efficiency trade-off of existing approaches for mitigating hallucinations in LVLMs and present the results in Figure~\ref{fig:teaser}. As we can see, while other hallucination mitigation methods achieve higher accuracy on the hallucination evaluation benchmark, they come at a significant cost, requiring 2$\times$ or more inference time and higher GPU memory consumption. We recognize this overhead is impractical given the limited performance improvements, highlighting the urgent need for more efficient approaches that can effectively mitigate hallucinations in LVLM.

In this work, we introduce ONLY, a training-free approach that requires only a single query and a one-layer intervention during decoding, offering an efficient solution for mitigating hallucinations in LVLMs. 
Our ONLY approach selects attention heads that prioritize textual information over visual information—specifically, those with a high text-to-visual entropy ratio—to stimulate textually enhanced next-token predictions. The enhanced output is then adaptively contrasted/collaborated with the original output logits using a single-layer intervention, aiming to reduce predominant and irrelevant language bias.
Our ONLY approach is both simple and effective, requiring just one additional attention layer computation. It incurs a modest 1.07$\times$ increase in inference time with negligible GPU memory overhead, significantly lower than the 2$\times$ or more increase seen in previous contrastive decoding methods. Moreover, ONLY achieves superior performance across multiple benchmarks, outperforming the current state-of-the-art by 3.14\% on POPE and 1.6\% on CHAIR.



To validate the effectiveness of our proposed ONLY approach, we evaluate it on three LVLMs (\textit{i.e.}, LLaVA-1.5~\citep{liu2024improved}, InstructBLIP~\citep{dai2024instructblip}, and Qwen-VL~\cite{bai2023qwen}) across various benchmarks, including POPE~\citep{li2023evaluating}, CHAIR~\citep{rohrbach2018object}, MME-Hallucination~\citep{fu2023mme}, MMBench~\cite{liu2025mmbench}, MM-Vet~\cite{yu2024mmvet}, and MMVP~\citep{tong2024eyes}. Extensive experimental results demonstrate that our ONLY approach consistently outperforms state-of-the-art methods across these benchmarks while requiring minimal implementation effort and computational cost. Additionally, qualitative case studies and GPT-4V-aided evaluations on LLaVA-Bench further validate the effectiveness of our ONLY approach in enhancing the coherence and accuracy of LVLM responses.

Our contributions are summarized as follows:
\begin{itemize}
    \item We investigate and challenge the performance-efficiency trade-off of existing contrastive decoding approaches for mitigating hallucinations in LVLMs, highlighting the efficiency issues.
    \item We present ONLY, a novel training-free decoding algorithm that leverages a single additional Transformer layer to improve the accuracy of LVLM responses.
    \item We conduct comprehensive experiments across various benchmarks and demonstrate that our proposed ONLY consistently outperforms existing approaches with minimal implementation effort and computational cost.
\end{itemize}

\section{Related Work}
\looseness=-1
\textbf{Large Vision-Language Models (LVLMs)}. 
Recently, large-scale LLMs have demonstrated remarkable proficiency in handling human queries and exhibit robust linguistic capabilities~\citep{touvron2023llama,chiang2023vicuna}. Leveraging these powerful models, researchers are exploring ways to align the visual modality with language, unlocking advanced visual recognition and reasoning capabilities across various multi-modal tasks~\citep{liu2024survey,bai2024hallucination}. For example, LLaVA-1.5~\citep{liu2023visual} employs a pre-trained CLIP ViT-L/14~\citep{radford2021learning} as the vision encoder, and trains a linear mapping layer to connect the vision and language modalities. In contrast, InstructBLIP~\citep{dai2024instructblip} builds on a pre-trained BLIP-2~\citep{li2023blip} and incorporates an instruction-aware Q-Former module to bridge the modalities.
Despite their exceptional multi-modal performance, these LVLMs still suffer from hallucinations, often generating text responses that do not accurately reflect the given image input~\citep{zhang2025selfcorrecting,zhangincorporating,chen2024detecting,chen2024alleviating,sun2023aligning}. Such hallucinations pose significant challenges for deploying these models in real-world applications. In this work, we propose a novel training-free algorithm that mitigates hallucinations while improving the efficiency of LVLMs for real-world deployment.

\looseness=-1
\textbf{Hallucination in LVLMs}. Recent studies have revealed that LVLMs may generate cross-modal inconsistencies between visual inputs and their corresponding responses, \textit{i.e.}, hallucinations, which can lead to misinformation and performance degradation~\citep{liu2024survey,leng2024mitigating}. To mitigate these hallucinations, early works have explored the use of additional robust instruction tuning on curated datasets~\citep{chen2024alleviating,sun2023aligning,jiang2024hallucination}. While effective, these methods require extensive and costly training, making them impractical for individual users. More recently, researchers have explored an alternative approach by developing variant methods based on contrastive decoding strategies, which mitigate hallucinations and enhance coherence by contrasting logits from counterpart outputs~\citep{favero2024multi,leng2024mitigating,chen2024halc}. However, we recognize that these methods require two or even multiple queries, which slows down LVLM response generation, making them less suitable for real-time applications. In response, we propose ONLY, a contrastive decoding-based approach that requires only a one-time query and a one-layer intervention during decoding, achieving competitive performance while effectively minimizing implementation efforts and computational costs.
\begin{figure*}[t]
  \begin{center}
     \makebox[\textwidth]{\includegraphics[width=\textwidth]{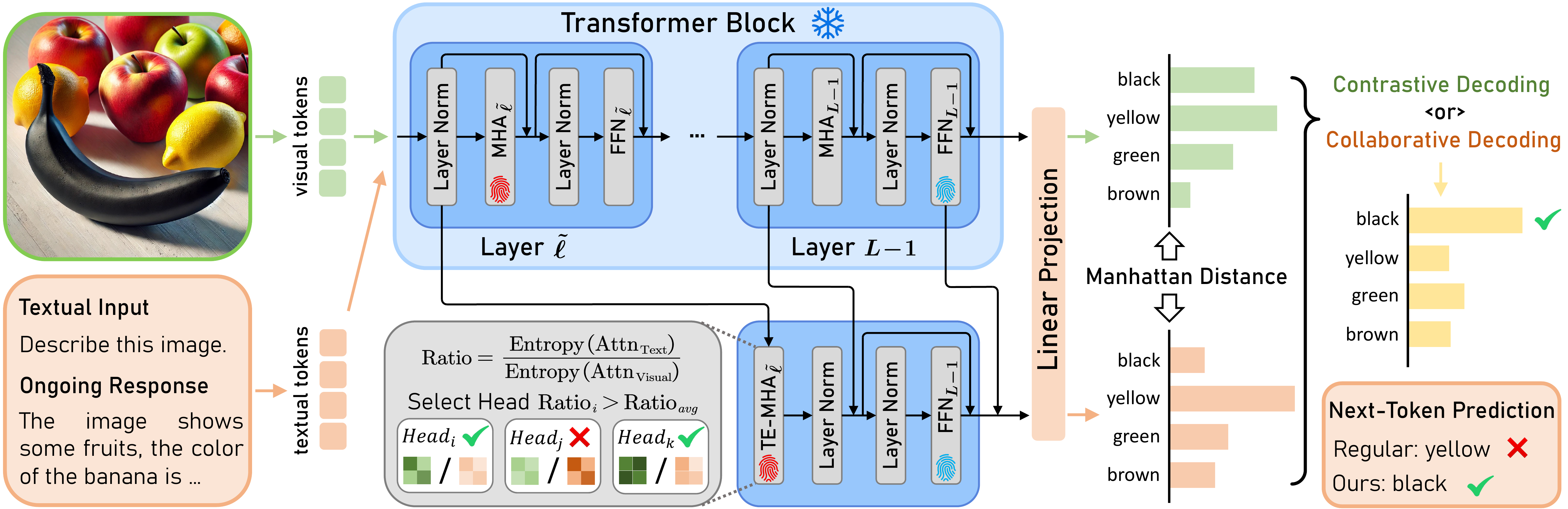}}
  \end{center}
  \vspace{-7pt}
  \caption{\textbf{Overview of our proposed ONLY}. Our method retains the core decoding process of LVLMs but incorporates a textual-enhanced multi-head attention layer with a residual connection to the last layer's output. This adjustment aims to produce an output with a greater focus on textual information. The resulting textual-enhanced logits are then adaptively decoded alongside the original output, employing either contrastive or collaborative decoding strategies to optimize performance.} 
  \vspace{-10pt}
  \label{fig:overview}
\end{figure*}

\section{Method}
\label{sec:method}
In this work, we present ONLY, a training-free algorithm that uses only one Transformer layer to improve the accuracy of LVLM responses, as illustrated in Figure~\ref{fig:overview}.
\subsection{Preliminaries}
\looseness=-1

\textbf{LVLM Decoding.} Recent LVLMs effectively process both visual and linguistic data using three key components: vision encoders, connectors, and a Large Language Model (LLM). An LVLM, parameterized by \( \theta \), autoregressively generates a fluent textual response sequence \( \mathbf{y} \) from an input image \( v \) and a textual query \( \mathbf{x} \). Initially, \( v \) is processed by a vision encoder and transformed into visual tokens via a vision-language alignment module (\textit{e.g.}, Q-Former~\citep{li2023blip} or linear projection~\citep{liu2023visual}). These tokens, combined with the query tokens, are input to the LLM. The generation of each token \( y_t \) in the sequence \( \mathbf{y} \) is modeled as:
\begin{equation}
\label{eq:regular decoding}
y_t \sim p_{\theta}(y_t | v, \mathbf{x}, \mathbf{y}_{<t}) = \text{softmax}(f_{\theta}(y_t | v, \mathbf{x}, \mathbf{y}_{<t}))_{y_t},
\end{equation}
where \( y_t \in \mathcal{S} \) is the current token, \( \mathbf{y}_{<t} = [y_0, \dots, y_{t-1}] \) are the previously generated tokens, and \( f_{\theta} \) represents the logits over a vocabulary set \( \mathcal{S} \).

\textbf{Transformer Decoder.} 
The language model is structured as a Transformer, where a sequence of tokens $\{x_1, x_2, \ldots, x_{t-1}\}$ are initially embedded into a sequence of hidden states $\mathcal{H}_{t-1}^0 = \{h_1^{0}, \ldots, h_{t-1}^{0}\}$. The Transformer comprises \(L\) layers, each layer incorporates a Multi-Head Attention (MHA) module and a Multi-Layer Perceptron (MLP). At time step \(t\), the output of each layer \( \mathcal{H}_t^{{\ell+1}} \) is derived from the hidden states input \( \mathcal{H}_t^{\ell} \), employing two primary residual connections:
\begin{equation}
\label{eq: attention output}
\bar{\mathcal{H}}_t^\ell = \text{MHA}_\ell(\mathcal{H}_t^{\ell}) + \mathcal{H}_t^{\ell}, \quad \mathcal{H}_t^{\ell+1} = \text{MLP}_\ell(\bar{\mathcal{H}}_t^\ell) + \bar{\mathcal{H}}_t^\ell.
\end{equation}
Each MHA module consists of \( H \) attention heads that compute self-attention, where the attention score is derived from query, key, and value matrices. Specifically, for the \(i\)-th head in layer \(\ell\), the operation is given by:
\begin{align}
\text{Head}_{\ell, i}(\mathcal{H}_t^{\ell}) &= \text{Attention}(Q_{\ell, i}, K_{\ell, i}, V_{\ell, i}) \nonumber\\ &= \text{softmax}\left(\frac{Q_{\ell, i}\cdot K_{\ell, i}^\top}{\sqrt{d_k}}\right) V_{\ell, i},
\end{align}
where \(Q_{\ell, i}/K_{\ell, i}/V_{\ell, i} = \mathcal{H}_t^{\ell}W^{Q/K/V}_{\ell, i}\), are query/key/value matrices obtained from learned weights. The outputs from all \(H\) heads are then concatenated and projected using an projection matrix \(W^O_\ell\):
\begin{align}
\text{MHA}_\ell(\mathcal{H}_t^{\ell}) = & \text{Concat}(\text{Head}_{\ell, 1}(\mathcal{H}_t^{\ell}), \nonumber \\&\text{Head}_{\ell, 2}(\mathcal{H}_t^{\ell}),  \ldots, \text{Head}_{\ell, H}(\mathcal{H}_t^{\ell}))W^O_\ell.
\end{align}
At last, a projection head $\phi(\cdot)$ predicts the logits of the next token $x_t$ over the vocabulary set \( \mathcal{S} \):
\begin{equation}
\label{eq:llm get logit}
f_{\theta}(y_t | y_{<t}) = \phi(\mathcal{H}_t^{L}), y_t \in \mathcal{S}.
\end{equation}
Combining Eq.~\ref{eq:llm get logit} with Eq.~\ref{eq:regular decoding}, we finally obtain:
\begin{align}
    p_{\theta}(y_t | v, \mathbf{x}, \mathbf{y}_{<t}) &= \text{softmax}(\phi(\mathcal{H}_t^{L}))_{y_t}.
\end{align}


\begin{algorithm}[t]
\small
\caption{Predict Textual-Enhanced (TE) Logits}
\label{alg:predict_contrastive_logits}
\begin{algorithmic}[1]
\Require Initial hidden states $\mathcal{H}_t^0$, total transformer layers $L$, total attention heads $H$, layer index for textual enhancement $\tilde{\ell}$.
\Procedure{predict\_te\_logits}{$A$}

    \For{$\ell \in \{0,1,2,\dots, L-1\}$} \hfill
        \For{$i \in \{0,1,\dots,H-1\}$}  \hfill
        \State \textcolor{maroonx}{{\textbf{Step 1:} Calculate TE attention output}}
            \If{$\ell = \tilde{\ell}$}
                \State
                \(
                \tilde{\mathcal{H}}_t^{\tilde{\ell}} \leftarrow \text{TE-MHA}_{\tilde{\ell}}(\mathcal{H}_t^{\tilde{\ell}})
                \) \Comment{Equation~\ref{Eq. TE-MHA}}
            \EndIf
        \EndFor

        \State \textcolor{maroonx}{{\textbf{Step 2:} Calculate Transformer output for each layer}}
        \State 
            \(
                \bar{\mathcal{H}}_t^{\ell} \leftarrow \text{MHA}_{\ell}(\mathcal{H}_t^{\ell}) + \mathcal{H}_t^{\ell}
            \) \Comment{Equation~\ref{eq: attention output}}
        \State
        \(
        \mathcal{H}_t^{\ell+1} \leftarrow \text{MLP}_{\ell}(\bar{\mathcal{H}}_t^{\ell}) + \bar{\mathcal{H}}_t^{\ell}
        \) \Comment{Equation~\ref{eq: attention output}}

        \If{$\ell = L-1$}
            \State \textcolor{maroonx}{{\textbf{Step 3:} Calculate TE Transformer output}}
            \State 
            \(
                \bar{\Tilde{\mathcal{H}}}_t^{L-1} \leftarrow \Tilde{\mathcal{H}}_t^{\tilde{\ell}} + \mathcal{H}_t^{L-1}
            \) \Comment{Equation~\ref{eq. te attention 1}}
            \State
            \(
                \hat{\mathcal{H}}_t^L \leftarrow \text{MLP}_{L-1}(\bar{\Tilde{\mathcal{H}}}_t^{L-1}) + \bar{\Tilde{\mathcal{H}}}_t^{L-1}
            \) \Comment{Equation~\ref{eq. te attention 2}}
        \EndIf
    \EndFor

    \State \textcolor{maroonx}{{\textbf{Step 4:} Calculate original logits and TE logits}}
    \State
    \(
        \mathrm{Logits} = f_{\theta}(y_t | v, \mathbf{x}, \mathbf{y}_{<t})) \leftarrow \texttt{Linear}(\mathcal{H}_t^{L})
    \)
    \State
    \(
        \mathrm{Logits\_TE} = \hat{f}_{\theta}(y_t | v, \mathbf{x}, \mathbf{y}_{<t})) \leftarrow \texttt{Linear}(\hat{\mathcal{H}}_t^L+\mathcal{H}_t^{L})
    \)

    \State \Return $\mathrm{Logits}$, $\mathrm{Logits\_TE}$ \hfill
\EndProcedure
\end{algorithmic}
\end{algorithm}

\begin{figure}[t]
  \centering
  \includegraphics[width=\linewidth]{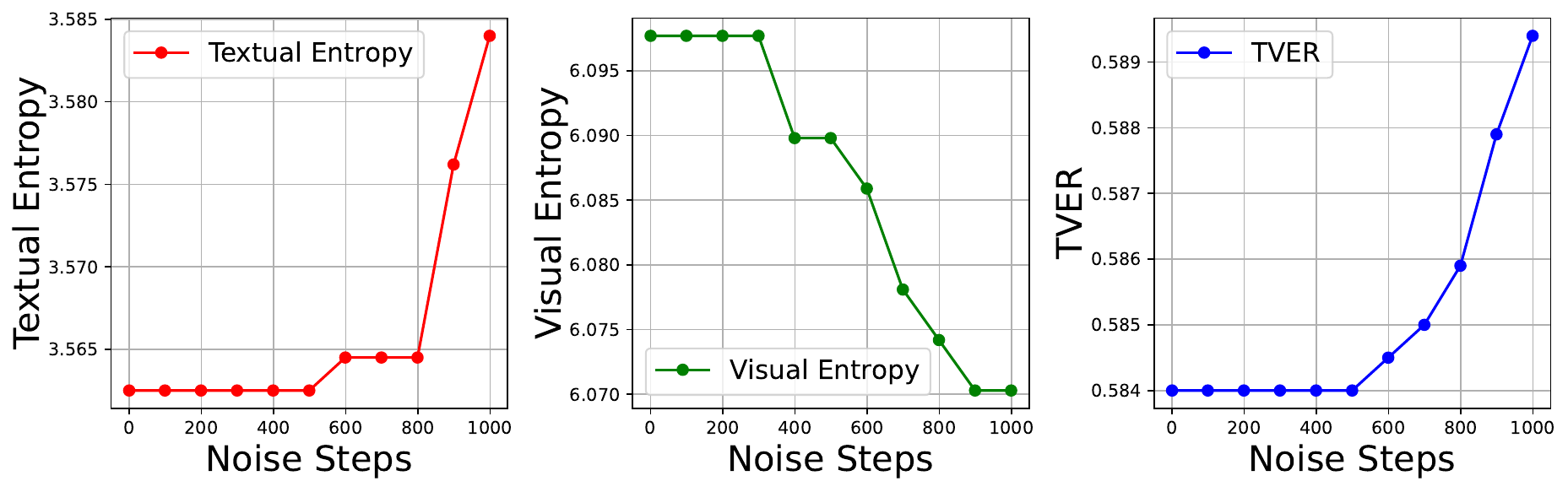}
  \includegraphics[width=\linewidth]{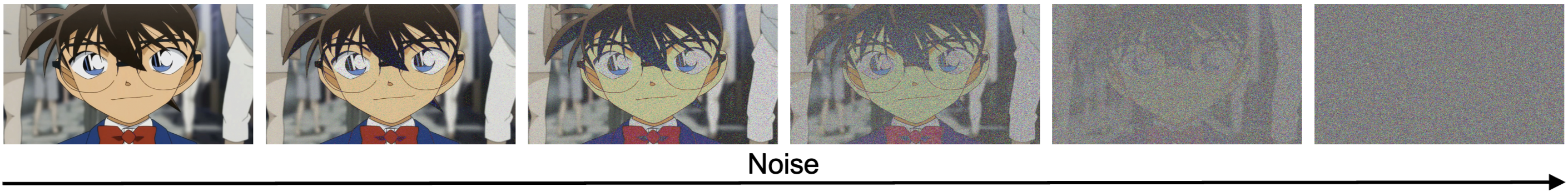}
  \vspace{-15pt}
  \caption{\looseness=-1\textbf{Impact of applying diffusion noise on textual and visual attention entropy}. We perform an analysis on all COCO samples from the POPE benchmark and observe that as distortion increases, textual entropy rises whereas visual entropy decreases.}
  \label{fig:tver}
  \vspace{-10pt}
\end{figure}


\subsection{One-Layer Intervention for Textual Enhancement}
Previous contrastive decoding methods focus primarily on the visual modality or the effect of visual input on the textual modality: \textit{e.g.}, VCD~\citep{leng2024mitigating} contrasts the outputs obtained with original vs. visual distorted input, and M3ID~\citep{favero2024multi} amplifies the influence of the reference image over the language prior. However, the effect of textual modality has been less studied.
In this work, we propose to address hallucination by directly producing textually-enhanced outputs with minimal additional computational overhead. Specifically, inspired by information theory~\citep{shannon1948mathematical}, we introduce an attention-head selection strategy guided by the text-to-visual entropy ratio. As illustrated in Figure~\ref{fig:tver}, we observe that when the distortion escalates (similar to the diffusion steps in VCD), textual entropy increases while visual entropy decreases. Guided by this observation, we propose to directly select attention heads with a higher text-to-visual entropy ratio to stimulate textually-enhanced outputs to avoid double queries while extracting language bias.
\textbf{Attention Head Selection Using Text-to-Visual Entropy Ratio.} 
Suppose a token is generated at time step \( t \), and the initial hidden states input to the Transformer decoder for this token is \( \mathcal{H}_t^0 \). For layer \( \ell \), the input hidden states can be denoted as \( \mathcal{H}_t^{\ell} \). We distinguish between text and visual attention within the attention matrix by computing the raw attention scores \( a_{\ell,i} \) for each head:
\begin{equation}
    a_{\ell,i} = \text{softmax}(Q_{\ell,i} \cdot K_{\ell,i}^\top/\sqrt{d_k}),
\end{equation}
where \( Q_{\ell,i} \) and \( K_{\ell,i} \) are the query and key matrices for head \( i \) in layer \( \ell \). To isolate text and visual attentions, we utilize indices corresponding to textual or visual tokens:
\begin{equation}
    \label{eq:texual/visual attention}
     a_{\ell,i}^\mathcal{T} = \{a_{\ell,i,j} \mid j \in \text{indices}_{\mathcal{T}}\}, a_{\ell,i}^\mathcal{V} = \{a_{\ell,i,j} \mid j \in \text{indices}_{\mathcal{V}}\},
\end{equation}
where \( \text{indices}_{\mathcal{T}} \) and \( \text{indices}_{\mathcal{V}} \) specify positions of textual and visual tokens, respectively. The entropy for these attention sets is computed as follows:
\begin{align}
    \text{Entropy}(a_{\ell,i}^\mathcal{T}) &= -\sum_{k} p_{\ell,i,k}^\mathcal{T} \log p_{\ell,i,k}^\mathcal{T}, \nonumber\\ \text{Entropy}(a_{\ell,i}^\mathcal{V})  &= -\sum_{k} p_{\ell,i,k}^\mathcal{V} \log p_{\ell,i,k}^\mathcal{V},
\end{align}
where \( p_{\ell,i,k}^\mathcal{T} \) and \( p_{\ell,i,k}^\mathcal{V} \) represent the normalized attention probabilities, computed from the softmax of each subset:
\begin{equation}
    p_{\ell,i,k}^\mathcal{T} = \text{softmax} (a_{\ell,i,k}^\mathcal{T}), p_{\ell,i,k}^\mathcal{V} = \text{softmax} (a_{\ell,i,k}^\mathcal{V}).
\end{equation}
The Text-to-Visual Entropy Ratio (\(\mathrm{TVER}\)) for each attention head is calculated as:
\begin{equation}
    \mathrm{TVER}_{\ell,i} = \frac{\text{Entropy}(a_{\ell,i}^\mathcal{T})}{\text{Entropy}(a_{\ell,i}^\mathcal{V})}.
\end{equation}
To optimize the attention output for enhanced textual relevance while reducing visual information, we selectively deactivate heads with a \(\mathrm{TVER}\) below the average for that layer, setting their attention weights to zero. 
This approach prioritizes heads with relatively higher text-to-visual entropy ratios, providing a clue where uncertainty in the textual modality is higher:
\begin{equation}
\label{eq:select head}
    \Tilde{a}_{\ell,i} = \begin{cases} 
    a_{\ell,i}, & \text{if } \mathrm{TVER}_{\ell,i} \geq \text{average}(\mathrm{TVER}_{\ell}), \\
    0, & \text{otherwise}.
    \end{cases}
\end{equation}
\looseness=-1
With this, we obtain the output of the Textual-Enhanced Multi-Head Attention (TE-MHA) module:
\begin{align}
\label{Eq. TE-MHA}
&\text{TE-MHA}_\ell(\mathcal{H}_t^{\ell}) = \nonumber\\ &\text{Concat}(\Tilde{a}_{\ell,1}V_{l,1}, \Tilde{a}_{\ell,2}V_{l,2}, \ldots, \Tilde{a}_{\ell,H}V_{l,H})W^O_\ell.
\end{align}

\subsection{Adaptive Decoding}
In this section, we utilize the logits obtained from textual-enhanced attention outputs for adaptive decoding.

\begin{figure}[t]
  \centering
  \includegraphics[width=\linewidth]{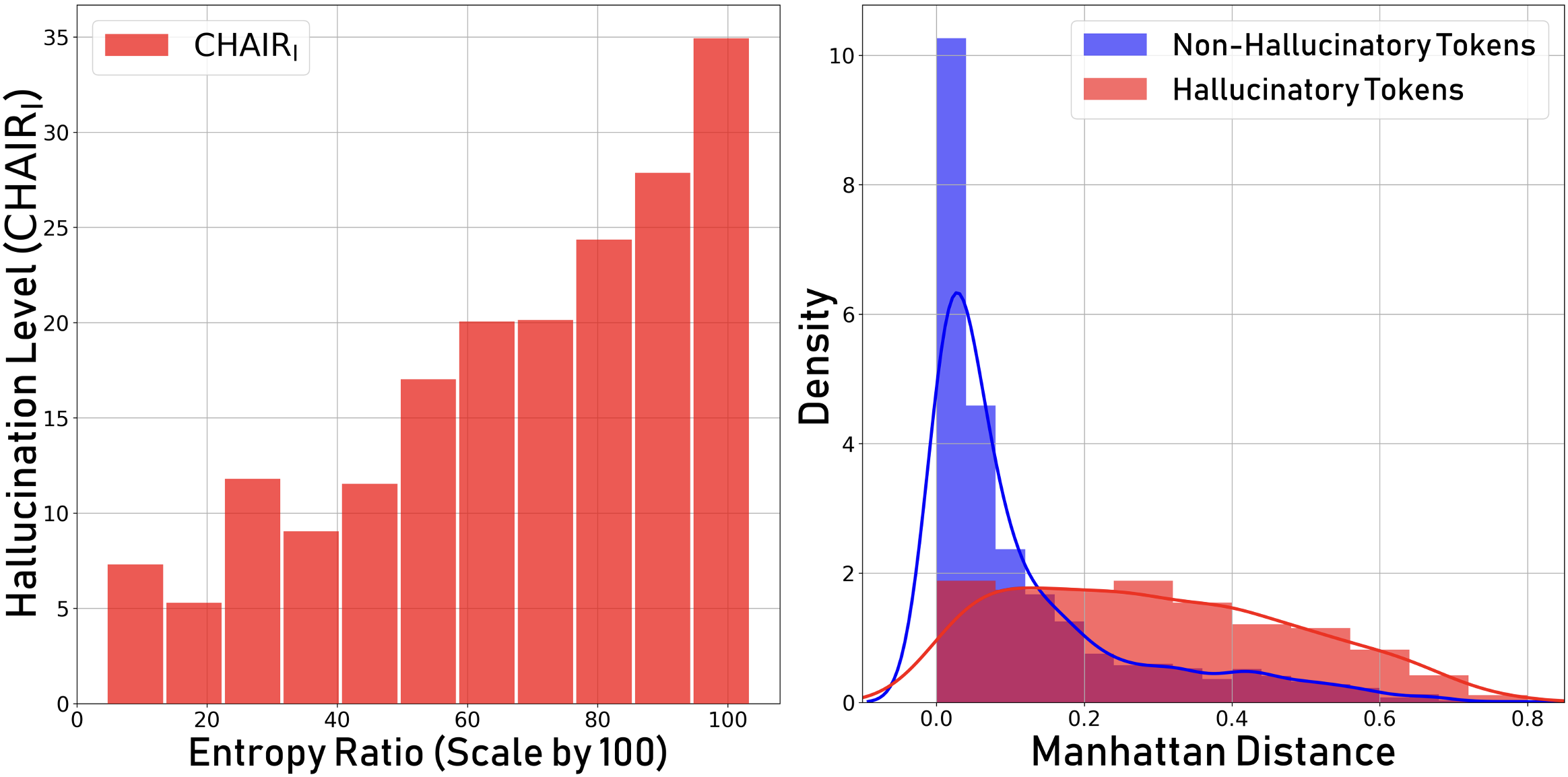}
  \vspace{-10pt}
  \caption{\textbf{Text-to-visual entropy ratio is correlated with hallucinations.} (\textit{Left}) Density plot of token-wise average textual-to-visual entropy ratio and bar plot of average $\text{CHAIR}_I$ in each bin on the CHAIR benchmark; (\textit{Right}) Density plots of token-level Manhattan distance between original and textual-enhanced logits for both hallucinatory and non-hallucinatory tokens on POPE.}
  \label{fig:feedback}
  \vspace{-10pt}
\end{figure}

Suppose layer $\Tilde{\ell} \in \{0, 1, \cdots, L-1\}$ is the selected layer for textual enhancement, where we calculate a textual-enhanced attention output as discussed in Eq.~\ref{Eq. TE-MHA}. To ensure that the output logits do not deviate excessively from the original LVLM outputs, we implement two residual connections. These connections are defined as follows:
\begin{align}
    &\Tilde{\mathcal{H}}_t^{\Tilde{\ell}} = \text{TE-MHA}_{\Tilde{\ell}}(\mathcal{H}_t^{\Tilde{\ell}}), \, \\
    \label{eq. te attention 1}
    &\bar{\Tilde{\mathcal{H}}}_t^{L-1} = \Tilde{\mathcal{H}}_t^{\Tilde{\ell}} + \mathcal{H}_t^{L-1}, \,  \\
    \label{eq. te attention 2}
    &\hat{\mathcal{H}}_t^L = \text{MLP}_{L-1}(\bar{\Tilde{\mathcal{H}}}_t^{L-1}) + \bar{\Tilde{\mathcal{H}}}_t^{L-1}.
\end{align}
Finally, the textual-enhanced predicted probability can be obtained by:
\begin{align}
    \Tilde{p}_{\theta}(y_t | v, \mathbf{x}, \mathbf{y}_{<t}) &= \text{softmax}(\Tilde{f}_{\theta}(y_t | v, \mathbf{x}, \mathbf{y}_{<t}))_{y_t} \nonumber\\&= \text{softmax}(\phi(\hat{\mathcal{H}}_t^{L}))_{y_t}.
\end{align}

To adaptively contrast the original and textual-enhanced logits, we measure the Manhattan distance between the two probability distributions at each timestep \(t\):
\begin{equation}
    d_t = \sum_{y_t \in \mathcal{S}} |p_{\theta}(y_t | v, \mathbf{x}, \mathbf{y}_{<t}) - \Tilde{p}_{\theta}(y_t | v, \mathbf{x}, \mathbf{y}_{<t})|,
\end{equation}
where \( d_t \) provides a measure of the difference between the distributions. Based on this distance, we adjust the original logits either collaboratively or contrastively:
\begin{equation}
y_t \sim p_{\theta}(y_t) =
\text{softmax} \left(f_\theta^{\textrm{final}}\right)
\end{equation}
\vspace{2pt}
\begin{equation}
f_\theta^{\textrm{final}} = 
\begin{cases} 
f_\theta(y_{t}| v,\mathbf{x},\mathbf{y}_{<t}) + \alpha_{1} \, \Tilde{f}_\theta(y_{t}| v,\mathbf{x},\mathbf{y}_{<t}), \\ \hfill\text{if } d_t < \gamma \text{ (collaborative)}; \\[5pt]
(1 + \alpha_{2}) \, f_\theta(y_{t}| v,\mathbf{x},\mathbf{y}_{<t}) - \alpha_{2} \, \Tilde{f}_\theta(y_{t}| v,\mathbf{x},\mathbf{y}_{<t}),  \\ \hfill\text{if } d_t \geq \gamma \text{ (contrastive)},
\end{cases}
\end{equation}
where \(\gamma\) is a predefined threshold that determines the decoding strategy based on the measured distance.

\textbf{Effectiveness of text-to-visual entropy ratio for textual information enhancement.} 
We further conduct an empirical study to validate the effectiveness of applying text-to-visual entropy ratio for language bias reflection, as shown in Figure~\ref{fig:feedback}.
The experimental results demonstrate that \textit{the entropy ratio is strongly correlated to the hallucination level} at both the response and token levels.


\section{Experiments}
\label{sec:experiment}
In this section, we evaluate the effectiveness of our method in mitigating hallucinations in LVLMs across a range of benchmarking scenarios, comparing it with existing state-of-the-art approaches.

\subsection{Experimental Settings}
\looseness=-1

\textbf{Evaluated LVLMs}. {We evaluate the effectiveness of our method on three state-of-the-art open-source LVLMs: LLaVA-1.5~\citep{liu2024improved}, InstructBLIP~\citep{dai2024instructblip} and Qwen-VL~\cite{bai2023qwen}.

\textbf{Benchmarks}. We conduct extensive experiments on six benchmarks: (1) \textbf{POPE~\citep{li2023evaluating}} is a benchmark commonly used to assess object hallucinations in LVLMs, which evaluates model accuracy through yes-or-no questions about the presence of specific objects in images; (2) \textbf{CHAIR~\citep{rohrbach2018object}} evaluates object hallucinations through image captioning, where the LVLMs are prompted to describe 500 randomly selected images from the MSCOCO validation set; (3) \textbf{MME-Hallucination~\citep{fu2023mme}} is a comprehensive benchmark for LVLMs consisting of four subsets: \textit{existence} and \textit{count} for object-level hallucinations, and \textit{position} and \textit{color} for attribute-level hallucinations; (4) {\textbf{MMBench~\citep{liu2025mmbench}} is a benchmark for evaluating LVLMs' multi-modal understanding ability across 20 dimensions}; (5) \textbf{MMVP~\citep{tong2024eyes}} comprises 150 CLIP-blind image pairs, each paired with a binary-option question to evaluate the fine-grained visual recognition capabilities of LVLMs; (6) \textbf{MM-Vet~\cite{yu2024mmvet}} utilizes LLM-based evaluator to evaluate LVLMs on 6 capabilities, including recognition, OCR, knowledge, language generation, spatial awareness, and math; (7) \textbf{LLaVA-Bench} provides 24 images in complex scenes, memes, and sketches, along with 60 challenging questions.

\begin{table*}[t]
    \centering
    \small
    \setlength{\tabcolsep}{5pt} 
    \resizebox{\textwidth}{!}{
    \begin{tabular}{cclcccccccccccc}
    \toprule
     & \multirow{2}{*}[-2pt]{{Setup}} & \multirow{2}{*}[-2pt]{{Method}} & \multicolumn{4}{c}{{LLaVA-1.5}} & \multicolumn{4}{c}{{InstructBLIP}} & \multicolumn{4}{c}{{Qwen-VL}} \\
    \arrayrulecolor{gray} \cmidrule(lr){4-7} \cmidrule(lr){8-11} \cmidrule(lr){12-15}
     &  &  & {Acc.} $\uparrow$ & {Prec.} $\uparrow$ & {Rec.} $\uparrow$ & {F1} $\uparrow$ & {Acc.} $\uparrow$ & {Prec.} $\uparrow$ & {Rec.} $\uparrow$ & {F1} $\uparrow$ & {Acc.} $\uparrow$ & {Prec.} $\uparrow$ & {Rec.} $\uparrow$ & {F1} $\uparrow$ \\
    \midrule
    \multirow{12}{*}[-5pt]{\rotatebox{90}{{\normalsize \textbf{MS-COCO}}}} & \multirow{4}{*}{Random} 
    & Regular & 83.13 & 81.94 & 85.00 & 83.44 & 83.07 & 83.02 & 83.13 & 83.08 & 85.23 & 97.23 & 72.53 & 83.09 \\
     &  & VCD  & 87.00 & 86.13 &  \underline{88.20} & 87.15 & 86.23 & \underline{88.14} & 83.73 & 85.88 & \underline{87.03} & 97.36 & \underline{76.13} & \underline{85.45} \\
     &  & M3ID  &  \underline{87.50} &  \underline{87.38} & 87.67 &  \underline{87.52} &  \underline{86.67} &  88.09 &  \underline{84.80} &  \underline{86.41} & 86.40 & \underline{98.23} & 74.13 & 84.50 \\
     &  & \cc {\textbf{Ours}} &\cc \textbf{89.70} &\cc \textbf{89.95} &\cc \textbf{88.27} &\cc \textbf{89.10} &\cc \textbf{89.23}  &\cc \textbf{91.83} &\cc \textbf{86.13} &\cc \textbf{88.89} & \cc \textbf{88.90} & \cc \textbf{98.52} & \cc \textbf{79.27} & \cc \textbf{87.85} \\
     \arrayrulecolor{gray}\cmidrule(lr){2-15}
      &  \multirow{4}{*}{Popular} & Regular & 81.17 &	78.28 &	86.27 &	82.08 & 77.00	& 73.82 &	83.67 &	78.44 & 84.53 & 94.50 & 73.33 & 82.58 \\
     &  & VCD  & 83.10 & 79.96 &  \underline{88.33} & 83.94 &  80.07 &  77.67 & 84.40 & 80.89 & 85.87 & 94.98 & \underline{75.73} & 84.27 \\
     &  & M3ID  &  \underline{84.30} &  \underline{81.58} &  \textbf{88.60} &  \underline{84.95} & \underline{80.97} & \underline{77.93} &  \textbf{{86.40}} &  \underline{81.85} & \underline{86.07} & \textbf{96.56} & 74.80 & \underline{84.30} \\
     &  & \cc {\textbf{Ours}} &\cc \textbf{86.00} &\cc \textbf{84.44} &\cc 88.27 &\cc \textbf{86.31} &\cc \textbf{83.27}  &\cc \textbf{81.46} &\cc \underline{86.13} &\cc \textbf{83.73} & \cc \textbf{87.47} & \cc \underline{95.63} & \cc \textbf{79.48} & \cc \textbf{86.81} \\
     \arrayrulecolor{gray}\cmidrule(lr){2-15}
      &  \multirow{4}{*}{Adversarial} & Regular & 77.43 & 73.31 & 86.27 & 79.26 & 74.60 & 71.26 & 82.47 & 76.45 & 83.37 & \underline{91.47} & 73.60 & 81.57 \\
     &  & VCD  & 77.17 & 72.18 & \textbf{88.40} & 79.47 &  77.20 &  \underline{74.29} & 83.20 &  78.49 & \underline{83.73} & 89.84 & \underline{76.07} & \underline{82.38} \\
     &  & M3ID  &  \underline{78.23} &  \underline{73.51} &  \underline{88.27} &  \underline{80.22} & \underline{77.47} & 73.68 &  \underline{85.47} & \underline{79.14} & 83.37 & 91.19 & 73.87 & 81.62 \\
     &  & \cc {\textbf{Ours}} &\cc \textbf{79.40} &\cc \textbf{75.00} &\cc 88.20 &\cc \textbf{81.07} &\cc \textbf{80.10}  &\cc \textbf{76.89} &\cc \textbf{86.07} &\cc \textbf{81.22} & \cc \textbf{83.80} & \cc \textbf{92.33} & \cc \textbf{76.14} & \cc \textbf{83.46} \\
     \arrayrulecolor{gray}\midrule
     \multirow{12}{*}[-5pt]{\rotatebox{90}{{\normalsize \textbf{A-OKVQA}}}} &  \multirow{4}{*}{Random} & Regular & 81.90 & 76.63 & 91.80 & 83.53 & 80.63 & 76.82 & 87.73 & 81.92 & 86.40 & 94.32 & 77.47 & 85.07 \\
     &  & VCD  & 83.83 & 78.05 & \underline{94.13} & 85.34 & 84.20 &  80.90 & 89.53 & 85.00 & \underline{87.93} & 94.59 & \underline{80.47} & \underline{86.96} \\
     &  & M3ID  &  \underline{84.67} &  \underline{79.25} &  93.93 &  \underline{85.97} &  \underline{85.43} & \underline{81.77} &  \underline{91.20} &  \underline{86.23} & 87.50 & \underline{95.33} & 78.87 & 86.32 \\
     &  & \cc {\textbf{Ours}} &\cc \textbf{86.07} &\cc \textbf{80.91} &\cc \textbf{94.40} &\cc \textbf{87.14} &\cc \textbf{88.57}  &\cc \textbf{86.13} &\cc \textbf{91.93} &\cc \textbf{88.94} & \cc \textbf{89.47} & \cc \textbf{95.34} & \cc \textbf{83.84} & \cc \textbf{89.22} \\
     \arrayrulecolor{gray}\cmidrule(lr){2-15}
      &  \multirow{4}{*}{Popular} & Regular & 75.07 &  68.58 & 92.53 & 78.77 & 75.17 & 70.15 & 87.60 & 77.91 & 85.77 & 92.82 & 77.53 & 84.49 \\
     &  & VCD  & 76.63 & 69.59 & \textbf{94.60} & 80.19 &  78.63 &  \underline{73.53} & 89.47 &  80.72 & 87.33 & 93.68 & \underline{80.07} & \underline{86.34} \\
     &  & M3ID  &  \underline{77.80} & \underline{70.98} & 94.07 & \underline{80.91} & \underline{78.80} & 73.38 & \underline{90.40} & \underline{81.00} & \underline{87.37} & \textbf{95.31} & 78.60 & 86.15 \\
     &  & \cc {\textbf{Ours}} &\cc \textbf{79.00} &\cc \textbf{72.17} &\cc \underline{94.40} &\cc \textbf{81.80} &\cc \textbf{80.83}  &\cc \textbf{75.23} &\cc \textbf{91.93} &\cc \textbf{82.75} & \cc \textbf{89.47} & \cc \underline{94.77} & \cc \textbf{84.43} & \cc \textbf{89.30} \\
     \arrayrulecolor{gray}\cmidrule(lr){2-15}
      &  \multirow{4}{*}{Adversarial} & Regular & 67.23 & 61.56 & 91.80 & 73.70 & 69.87 & 64.54 & 88.20 & 74.54 & 80.37 & 82.56 & 77.00 & 79.68 \\
     &  & VCD  & 67.40 & 61.39 & 93.80 & 74.21 &  \underline{71.00} &  \underline{65.41} &  89.13 &  \underline{75.45} & \underline{81.90} & 83.07 & \underline{80.13} & \underline{81.57} \\
     &  & M3ID  &  \underline{68.60} &  \underline{62.22} &  \textbf{94.73} &  \underline{75.11} & 70.10 &  64.28 & \underline{90.47} & 75.16 & \underline{81.90} & \underline{84.25} & 78.47 & 81.26 \\
     &  & \cc {\textbf{Ours}} &\cc \textbf{68.70} &\cc \textbf{62.35} &\cc \underline{94.40} &\cc \textbf{75.70} &\cc \textbf{72.47}  &\cc \textbf{66.19} &\cc \textbf{91.87} &\cc \textbf{76.94} & \cc \textbf{82.07} & \cc \textbf{85.02} & \cc \textbf{81.09} & \cc \textbf{83.01} \\
     \arrayrulecolor{gray}\midrule
     \multirow{12}{*}[-5pt]{\rotatebox{90}{{\normalsize \textbf{GQA}}}} &  \multirow{4}{*}{Random} & Regular & 82.23 & 76.32 & 93.47 & 84.03 & 79.67 & 76.05 & 86.60 & 80.99 & 85.10 & 91.42 & 77.47 & 83.87 \\
     &  & VCD  & 83.23 & 76.73 &  \underline{95.40} & 85.05 &  82.83 &  \underline{80.16} & 87.27 &  83.56 & 87.00 & 92.11 & \underline{80.93} & \underline{86.16} \\
     &  & M3ID  &  \underline{84.20} &  \underline{78.00} & 95.27 &  \underline{85.77} & \underline{83.07} & 80.06 & \underline{88.07} & \underline{83.87} & \underline{87.07} & \underline{92.64} & 80.53 & \underline{86.16} \\
     &  & \cc {\textbf{Ours}} &\cc \textbf{86.70} &\cc \textbf{80.94} &\cc \textbf{96.00} &\cc \textbf{87.83} &\cc \textbf{86.17}  &\cc \textbf{83.84} &\cc \textbf{89.60} &\cc \textbf{86.63} & \cc \textbf{88.03} & \cc \textbf{93.59} & \cc \textbf{82.68} & \cc \textbf{87.80}  \\
     \arrayrulecolor{gray}\cmidrule(lr){2-15}
      &  \multirow{4}{*}{Popular} & Regular & 73.47 & \textbf{66.83} & 93.20 & 77.84 & 73.33 & 68.72 & 85.67 & 76.26 & 80.87 & 82.65 & 78.13 & 80.33 \\
     &  & VCD  & 72.37 & 65.27 &  \underline{95.60} & 77.58 & \underline{76.13} & \underline{71.10} & 88.07 & \underline{78.68} & 82.53 & 83.52 & \underline{81.07} & \underline{82.27} \\
     &  & M3ID  &  \underline{73.87} &  \underline{66.70} & 95.33 &  \underline{78.49} &  75.17 &  69.94 & \underline{88.27} &  78.04 & \underline{82.68} & \underline{83.74} & 80.85 & \underline{82.27} \\
     &  & \cc {\textbf{Ours}} &\cc \textbf{74.03} &\cc \underline{66.70} &\cc \textbf{96.00} &\cc \textbf{78.71} &\cc \textbf{77.20}  &\cc \textbf{71.79} &\cc \textbf{89.60} &\cc \textbf{79.72} & \cc \textbf{82.87} & \cc \textbf{83.88} & \cc \textbf{82.55} & \cc \textbf{83.21} \\
     \arrayrulecolor{gray}\cmidrule(lr){2-15}
      &  \multirow{4}{*}{Adversarial} & Regular & 68.60 & \underline{62.43} & 93.40 & 74.84 &  68.60 &  63.94 & 85.33 &  73.10 & 78.77 & 79.33 & 77.80 & 78.56 \\
     &  & VCD  & \underline{68.83} & 62.26 & \underline{95.67} & \underline{75.43} & 71.00 & 65.75 & 87.67 & {75.14}& 81.17 & 81.48 & \underline{80.67} & 81.07 \\
     &  & M3ID  &  68.67 &  62.16 &  95.40 &  75.28 & \underline{71.17} & \underline{65.79} &  \textbf{88.20} & \underline{75.36} & \textbf{81.90} & \textbf{83.07} & 80.13 & \underline{81.57} \\
     &  & \cc {\textbf{Ours}} &\cc \textbf{69.23} &\cc \textbf{62.55} &\cc \textbf{95.87} &\cc \textbf{75.70} &\cc \textbf{71.93}  &\cc \textbf{65.98} &\cc \underline{87.93} &\cc \textbf{75.84} & \cc \underline{81.33} & \cc \underline{82.38} & \cc \textbf{81.50} & \cc \textbf{81.94} \\
    \bottomrule
    \end{tabular}
    }
    \vspace{-5pt}
    \caption{
        \textbf{Results on POPE~\citep{li2023evaluating} benchmark.} Higher ($\uparrow$) accuracy, precision, recall, and F1 indicate better performance. The best results are \textbf{bolded}, and the second-best are \underline{underlined}.
    }
    \label{tab:POPE}
\end{table*}

\begin{table*}[t]
        \begin{center}
        \begin{small}
        \setlength{\tabcolsep}{2pt} 
        \resizebox{\textwidth}{!}{
        \begin{tabular}{lcccccccccccc}
            \toprule
              \multirow{3}{*}[-0.5ex]{\textbf{Method}}  &  \multicolumn{4}{c}{\textbf{LLaVA-1.5}} & \multicolumn{4}{c}{\textbf{InstructBLIP}} & \multicolumn{4}{c}{\textbf{Qwen-VL}} \\
            \cmidrule(lr){2-5}\cmidrule(lr){6-9}\cmidrule(lr){10-13}
                 & \multicolumn{2}{c}{Max Token 64} & \multicolumn{2}{c}{Max Token 128} & \multicolumn{2}{c}{Max Token 64} & \multicolumn{2}{c}{Max Token 128} & \multicolumn{2}{c}{Max Token 64} & \multicolumn{2}{c}{Max Token 128} \\
             \cmidrule(lr){2-3}\cmidrule(lr){4-5}\cmidrule(lr){6-7}\cmidrule(lr){8-9}\cmidrule(lr){10-11}\cmidrule(lr){12-13}
                 & CHAIR$_S$ $\downarrow$ & CHAIR$_I$ $\downarrow$ & CHAIR$_S$ $\downarrow$ & CHAIR$_I$ $\downarrow$ & CHAIR$_S$ $\downarrow$ & CHAIR$_I$ $\downarrow$ & CHAIR$_S$ $\downarrow$ & CHAIR$_I$ $\downarrow$ & CHAIR$_S$ $\downarrow$ & CHAIR$_I$ $\downarrow$ & CHAIR$_S$ $\downarrow$ & CHAIR$_I$ $\downarrow$ \\
             \midrule
             Regular & 26.2 & 9.4 & 55.0 & 16.3 &  31.2 & 11.1  & 57.0 & 17.6 & 33.6 & 12.9 & 52.0 & 16.5 \\
              VCD & 24.4 & 7.9 & 54.4 & 16.6 &  {30.0} & {10.1}  & 60.4 & 17.8 & 33.0 & 12.8 & 50.2 & 16.8 \\
              M3ID & \underline{21.4} & \underline{6.3}  & 56.6 & 15.7 & 30.8 & 10.4 & 62.2 & 18.1 & 32.2 & 11.5 & \underline{49.5} & 17.2 \\
              {Woodpecker} & {24.9} & {7.5}  & 57.6 & 16.7 &  31.2 & 10.8 & 60.8 & 17.6 & 31.1 & 12.3 & 51.8 & 16.3 \\
              {HALC} & {21.7} & {7.1}  & \underline{51.0} & \underline{14.8} &  \underline{24.5} & \textbf{8.0} & \underline{53.8} & \underline{15.7} & \underline{28.2} & \underline{9.1} & {49.6} & \underline{15.4} \\
              \cc \textbf{Ours} & \cc \textbf{20.0} & \cc \textbf{6.2} & \cc \textbf{49.8} & \cc \textbf{14.3} & \cc \textbf{23.5} & \cc \underline{8.2} & \cc \textbf{52.2} & \cc \textbf{15.5} & \cc \textbf{27.3} & \cc \textbf{8.4} & \cc \textbf{48.0} & \cc \textbf{14.3} \\
            \bottomrule
        \end{tabular}
        }
        \vspace{-5pt}
        \caption{\textbf{Results on CHAIR~\citep{rohrbach2018object} benchmark.} We limit the maximum number of new tokens to 64 or 128. Lower ($\downarrow$) CHAIR$_S$, CHAIR$_I$ indicate better performance. The best results in each setting are \textbf{bolded}, and the second-best are \underline{underlined}.}
        \label{tab:CHAIR}
        \end{small}
        \end{center}
        \vspace{-20pt}
\end{table*}

\begin{table}[t!]
    \begin{center}
        \begin{small}
        \resizebox{\linewidth}{!}{
        \begin{tabular}{lccccc}
            \toprule
             \multirow{2}{*}[-0.5ex]{\textbf{Method}}  &  \multicolumn{2}{c}{\textbf{Object-level}} & \multicolumn{2}{c}{\textbf{Attribute-level}} & \multirow{2}{*}[-0.5ex]{\textbf{MME Score $\uparrow$}} \\
             \cmidrule(lr){2-3}\cmidrule(lr){4-5}
                & Existence $\uparrow$ & Count $\uparrow$ & Position $\uparrow$ & Color $\uparrow$   & \\
             \midrule
             Regular &  173.75 & 121.67 & 117.92 & 149.17 &  562.50 \\
              DoLa  & 176.67 & 113.33 & 90.55 & 141.67 &  522.22 \\
              OPERA  & 183.33 & \underline{137.22} & 122.78 & \underline{155.00} &  \underline{598.33} \\
              VCD  & 186.67 & 125.56 & 128.89 & 139.45 &  580.56 \\
              M3ID  & 186.67 &  128.33 & \underline{131.67} & 151.67 &  598.11 \\
              Woodpecker & \underline{187.50} & 125.00 & 126.66 & 149.17 &  588.33 \\
              HALC & 183.33 & 133.33 & 107.92 & \underline{155.00} &  579.58 \\
             \cc \textbf{Ours} & \cc \textbf{191.67} & \cc \textbf{145.55} & \cc \textbf{136.66} & \cc \textbf{161.66} & \cc \textbf{635.55} \\
            \bottomrule
        \end{tabular}
        }
        \vspace{-5pt}
        \caption{\looseness=-1\textbf{Results on MME-Hallucination~\citep{fu2023mme}  with LLaVA-1.5~\cite{liu2024improved}.} We report the average MME scores for each subset. Higher scores ($\uparrow$) indicate better performance. The best results are \textbf{bolded}, and the second-best are \underline{underlined}.}
        \label{tab:MME}
        \end{small}
        \vspace{-20pt}
    \end{center}
\end{table}

\textbf{Baselines}. We compare the performance of our ONLY approach with the following state-of-the-art approaches: VCD~\citep{leng2024mitigating}, M3ID~\citep{favero2024multi}, Woodpecker~\citep{yin2023woodpecker}, HALC~\citep{chen2024halc}, DoLa~\citep{chuang2024dola} and OPERA~\citep{huang2024opera}. We apply sampling-based decoding in default, where the next token is sampled directly from the post-softmax probability distribution.


\looseness=-1
\textbf{Implementation Details}. We follow the default query format for all LVLMs. Besides, we set $\alpha_1 = 3$, $\alpha_2 = 1$, and $\gamma = 0.2$ for LLaVA-1.5~\citep{liu2024improved}, and $\gamma = 0.4$ for InstructBLIP~\citep{dai2024instructblip} / Qwen-VL~\cite{bai2023qwen}. Following VCD~\citep{leng2024mitigating}, we implement adaptive plausibility constraints~\citep{li2023contrastive} with $\beta=0.1$ across all tasks. All experiments are performed on a single 48GB NVIDIA RTX 6000 Ada GPU.


\subsection{Results and Discussions}
\looseness=-1
\textbf{Results on POPE}. In Table~\ref{tab:POPE}, we compare our method’s performance against various baselines on the POPE benchmark. As shown in the table, our approach consistently outperforms previous state-of-the-art methods across various LVLM models and settings, demonstrating its robustness across different evaluation scenarios.
Specifically, in the MS-COCO (Random) setting with the LLaVA-1.5 backbone, our method surpasses VCD by 2.20\% and M3ID by 1.70\% in accuracy. Even in the more challenging adversarial setting, our approach maintains its superior performance, outperforming VCD by 2.23\% and M3ID by 1.17\%.
Overall, these consistent gains across different datasets and LVLM models highlight the effectiveness of our method as a strong and generalizable solution for mitigating hallucinations in LVLMs.

\textbf{Results on CHAIR}. On the open-ended CHAIR benchmark, our ONLY method achieves superior performance with lower hallucination rates. Table~\ref{tab:CHAIR} presents a comparison against four state-of-the-art approaches, evaluating hallucination rates with CHAIR$_S$ and CHAIR$_I$ under maximum token generation limits of 64 and 128 across three LVLM backbones. Notably, in the LLaVA-1.5 (Max Token = 128) setting, our approach reduces CHAIR$_S$ by 5.2 points and CHAIR$_I$ by 2.0 points compared to regular decoding.

\textbf{Results on MME}. In Table~\ref{tab:MME}, we compare our approach against other methods on the MME benchmark. The results show that our method consistently outperforms all baselines, achieving the highest scores across both object-level (Existence, Count) and attribute-level (Position, Color) evaluations.
Notably, our method attains an MME score of 634.67, outperforming the second-best method, M3ID, by 36.34 points, demonstrating its superior capability in mitigating various types of hallucinations.


\textbf{Results on MMVP}. To evaluate the effectiveness of our approach on fine-grained visual recognition tasks, we conduct experiments on the MMVP benchmark and present the results in Figure~\ref{fig:mmvp}. With our ONLY approach, the LVLM is able to handle more nuanced visual recognition tasks, improving the performance from 22.67\% to 28.00\%. 

\textbf{Results on MMBench and MMVet}. We also report the performance of all compared methods on the MMBench and MMVet benchmarks in Table~\ref{tab:efficiency}. Our approach continues to outperform existing state-of-the-art methods, demonstrating that it also enhances the general multi-modal understanding capabilities of LVLMs.

\begin{figure}[t]
    \centering
    \includegraphics[width=0.95\linewidth]{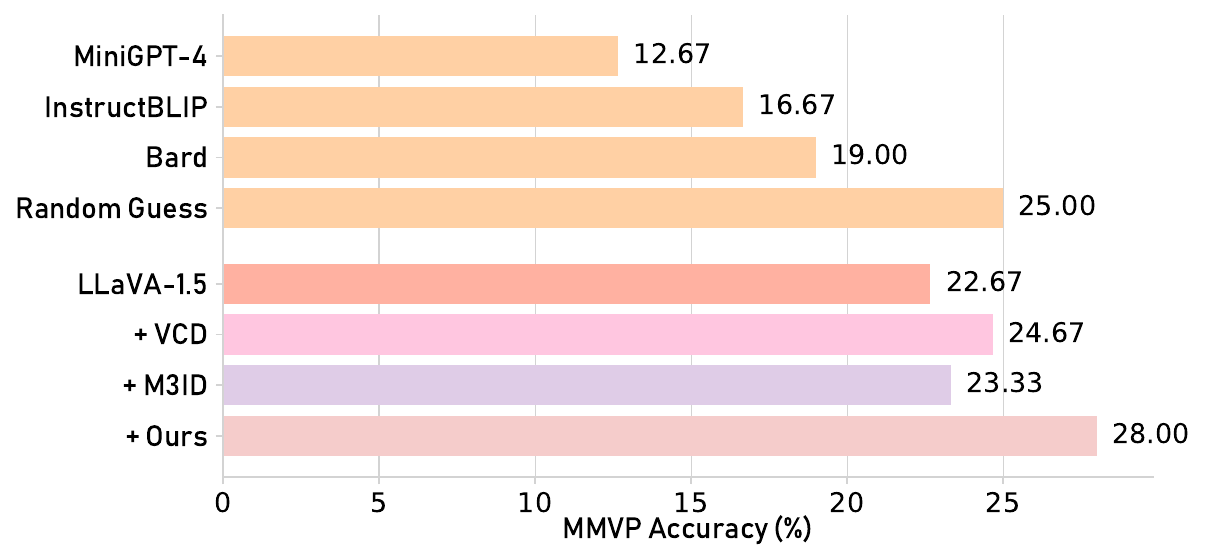}
    \vspace{-10pt}
    \caption{\looseness=-1\textbf{Results on MMVP~\citep{tong2024eyes}}. We apply our approach to LLaVA-1.5~\citep{liu2024improved} and compare its performance against other hallucination mitigation methods.}
    \label{fig:mmvp}
    \vspace{-8pt}
\end{figure}

\begin{table}[t]
    \centering
    \small
    \setlength{\tabcolsep}{10pt} 
    \resizebox{0.8\linewidth}{!}{
        \begin{tabular}{lcccc}
            \toprule
              \multirow{2}{*}[-0.5ex]{\textbf{Method}}  &  \multicolumn{2}{c}{\textbf{LLaVA-1.5}} & \multicolumn{2}{c}{\textbf{InstructBLIP}} \\
             \cmidrule(lr){2-3}\cmidrule(lr){4-5}
                 & Acc. $\uparrow$ & Det. $\uparrow$ & Acc. $\uparrow$ & Det. $\uparrow$ \\
             \midrule
             Regular & 6.07 & 6.20 & 5.26 & 5.53 \\
             \cc \textbf{Ours} & \cc \textbf{7.00} & \cc \textbf{7.13}  & \cc \textbf{6.60} & \cc \textbf{6.73} \\
             \midrule
              VCD & 4.60 & 5.13 & 4.87 & 5.33 \\
             \cc \textbf{Ours} & \cc \textbf{6.27} & \cc \textbf{6.60}  & \cc \textbf{6.80} & \cc \textbf{6.93} \\
              \midrule
              M3ID & 6.13 & 6.27 & 5.93 & 6.20 \\
             \cc \textbf{Ours} & \cc \textbf{6.73} & \cc \textbf{6.80}  & \cc \textbf{6.67} & \cc \textbf{6.87} \\
            \bottomrule
        \end{tabular}
    }
    \vspace{-5pt}
    \caption{\textbf{GPT-4V-aided evaluation on LLaVA-Bench}. Higher accuracy and detailedness ($\uparrow$) indicate better performance. The evaluation is performed on LLaVA-1.5~\citep{liu2024improved}.}
    \vspace{-15pt}
    \label{table:gpt4v}
\end{table}

\begin{table*}[t]
\small
\centering
\setlength{\tabcolsep}{9pt}
\resizebox{\linewidth}{!}{
\begin{tabular}{lccccccccc}
\toprule
Method  & Avg. Latency $\downarrow$ & GPU Memory $\downarrow$ & CHAIR$_S$ $\downarrow$ & MME $\uparrow$ & POPE $\uparrow$ & MMBench $\uparrow$ & MM-Vet $\uparrow$ \\ 
\midrule
Regular  &  3.47 s {\tiny ($\times$1.00)} & 14945 MB  {\tiny ($\times$1.00)} & 55.0 & 562.5 & 83.44 & 64.1 & 26.1 \\
VCD &  6.97 s {\tiny ($\times$2.01)} &  15749 MB  {\tiny ($\times$1.05)} & 54.4 & 580.6 & 87.15 & \underline{64.6} & 30.9 \\
M3ID &  7.05 s {\tiny ($\times$2.03)} &  15575 MB  {\tiny ($\times$1.04)} & 54.4 & 598.1 & 87.52 & 64.4 & 29.9 \\
OPERA & 24.70 s  {\tiny ($\times$7.12)} &  22706 MB  {\tiny ($\times$1.52)} & 52.6 & \underline{598.3} & \underline{88.85} & 64.4 & \underline{32.0} \\
Woodpecker & 10.68 s {\tiny ($\times$3.08)} & 22199 MB  {\tiny ($\times$1.49)} & 57.6 & 588.3 & 86.45 & 64.0 & 30.6 \\
HALC & 22.61 s {\tiny ($\times$6.52)} &  23084 MB {\tiny ($\times$1.54)} & \underline{51.0} & 579.6 & 87.68 & 64.2 & 30.8 \\
\rowcolor{gray!20}
\textbf{Ours} &  3.70 s {\tiny ($\times$1.07)}  & 14951 MB  {\tiny ($\times$1.00)} & \textbf{49.8} & \textbf{635.6} & \textbf{89.10} & \textbf{65.0} & \textbf{32.8} \\
\bottomrule
\end{tabular}
}
\vspace{-5pt}
\caption{{\textbf{Efficiency comparison}. For each method, we present the average inference latency per instance and peak GPU memory. Experiments are conducted on a single RTX A6000 Ada GPU.}}
\vspace{-10pt}
\label{tab:efficiency}
\end{table*}

\textbf{Results on LLaVA-Bench}. In Figure~\ref{fig:llavabench}, we present a case study on LLaVA-Bench comparing our method's response with the response generated by regular decoding using the LLaVA-1.5 model. Specifically, regular decoding often leads to hallucinated or inaccurate content, such as describing ``\texttt{taxi appears to be converted laundry machines}'' and ``\texttt{another person can be seen standing nearby}''. In contrast, our response is more detailed, focusing on the fact that ``\texttt{a person is ironing clothes while on the move, which is an unconventional way}''. The GPT-4V-aided evaluation in Table~\ref{table:gpt4v} further validates that our method improves both the accuracy and detailedness of generated responses.

\begin{figure}[t]
\centering
\includegraphics[width=\linewidth]{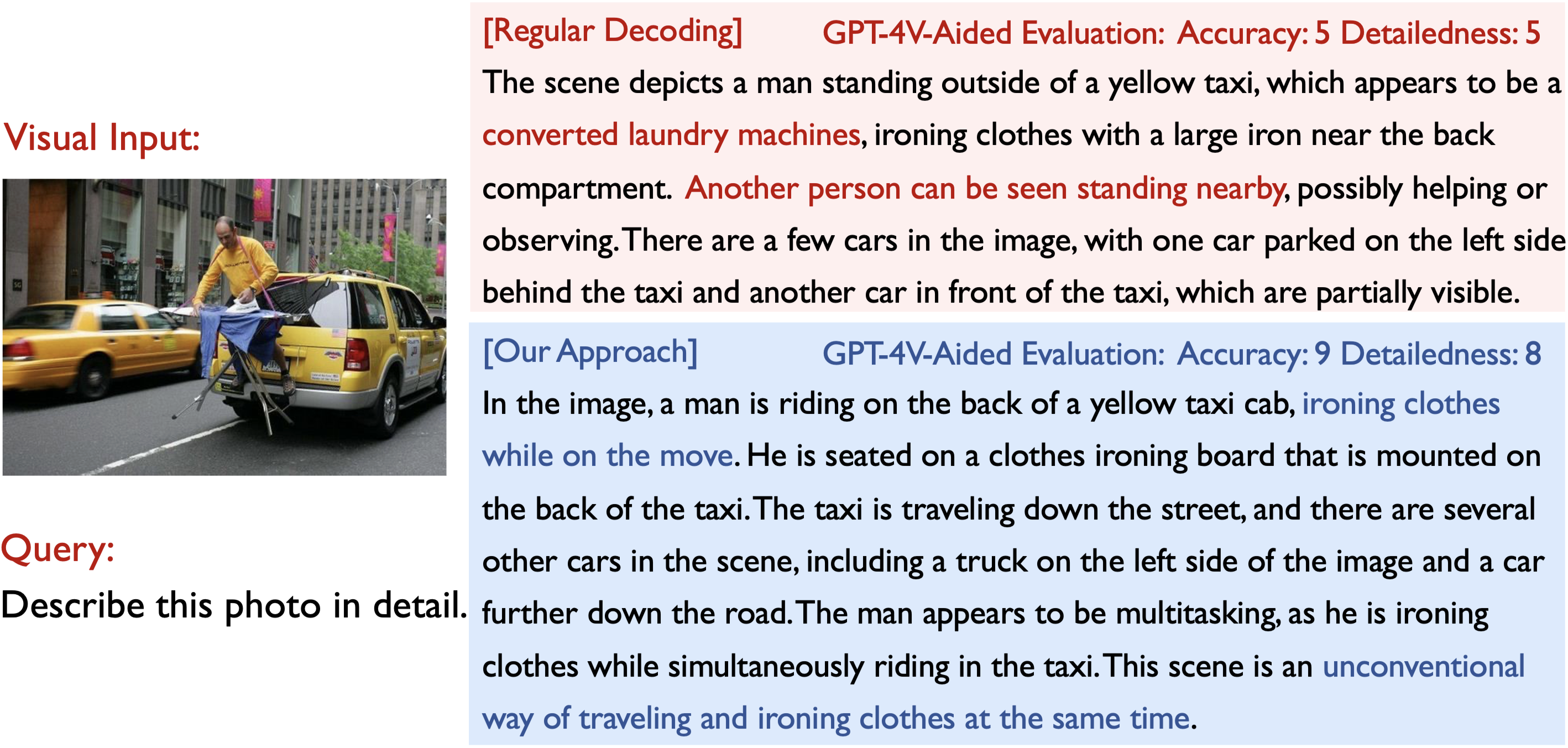}
\vspace{-18pt}
\caption{\textbf{Case study on the LLaVA-Bench benchmark}. We compare the responses generated by regular decoding and our method using LLaVA-1.5. GPT-4V-aided evaluation results are also provided alongside the responses. Hallucinated and accurate content is highlighted in \textcolor{darkred}{red} and \textcolor{darkblue}{blue}.} 
\vspace{-12pt}
\label{fig:llavabench}
\end{figure}

\subsection{Efficiency Comparison}
In Table~\ref{tab:efficiency}, we evaluate the efficiency of our approach using the LLaVA-1.5 model on the CHAIR benchmark, with a maximum token length of 128. We also report the performance of all compared methods across 5 benchmarks. Our approach demonstrates consistently superior performance, with only a 1.07$\times$ increase in time consumption and negligible additional GPU memory usage. These results validate that our approach is both efficient and effective, offering a favorable performance-cost trade-off.


\subsection{Ablation Study}
\label{sec:ablation on layer}
\textbf{Selection of Layer for Textual Enhancement.} To investigate the impacts of choosing different layers for textual enhancement, we conduct ablation experiments on the POPE benchmark.
Results in Figure~\ref{fig:layer} demonstrate that by selecting the initial layer for textual enhancement, our ONLY method achieves optimal performance on the POPE benchmark. Additionally, we observe that the performance of our approach is robust across different layers chosen for intervention, with ONLY exhibiting minimal variation and consistently outperforming VCD and M3ID. This robustness is due to our attention-head selection strategy, which dynamically selects different sets of heads across multiple layers, efficiently and effectively capturing language bias.

\begin{figure}[h]
    \centering
    \includegraphics[width=0.95\linewidth]{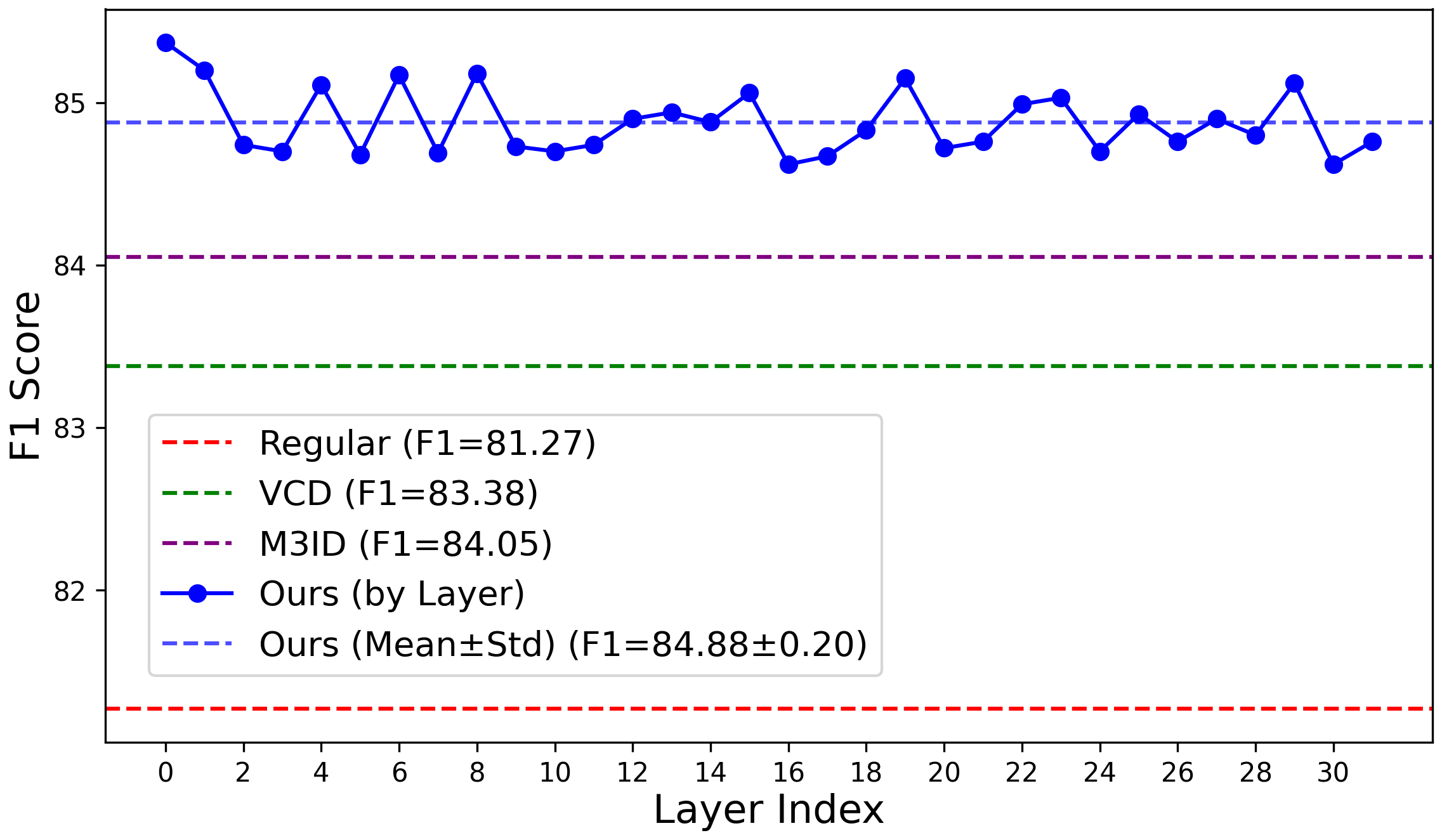}
    \vspace{-7pt}
    \caption{\looseness=-1\textbf{Impacts of different selected layers}. We present the results obtained by selecting different layers for textual enhancement on the POPE benchmark using all 9,000 samples from COCO.}
    \label{fig:layer}
\end{figure}

\textbf{Other Strategies for Textual Enhancement}.
In Table~\ref{tab:different strategies}, we compare the performance achieved by various textual enhancement strategies. Our approach of attention head selection using $\mathrm{TVER}$ achieves the best performance. In contrast, directly modifying attention weights—such as zeroing out or adding noise to visual attention weights, or doubling textual attention weights—results in suboptimal outcomes. Additionally, selecting attention heads based on the ratio of the sum of attention weights also leads to a performance decrease of 0.71\% on POPE and 3.1\% on CHAIR.

\begin{table}[t]
\centering
    \renewcommand{\arraystretch}{1.2}
    \resizebox{\linewidth}{!}{
    \begin{tabular}{lcccccccc}
            \toprule
              \multirow{2}{*}[-0.5ex]{\textbf{Strategy}}  &  \multicolumn{4}{c}{\textbf{POPE} $\uparrow$} & \multicolumn{2}{c}{\textbf{CHAIR} $\downarrow$} \\
             \cmidrule(lr){2-5}\cmidrule(lr){6-7}
                 & {Acc.}  & {Prec.}  & {Rec.}  & {F1}  & CHAIR$_S$  & CHAIR$_I$ \\
             \midrule
             Regular & 80.42 & 78.20 & 84.59 & 81.27 & 26.2  & 9.4 \\
             $a_{\ell,i}^\mathcal{V} \leftarrow 0$ & 84.26 & 82.13 & 87.69 & 84.82 & 21.2 & 6.9 \\
             $a_{\ell,i}^\mathcal{V} \leftarrow a_{\ell,i}^\mathcal{V}+\varepsilon$ & 83.95 & 81.67 & 88.16 & 84.79 & 22.1 & 7.6 \\
             $a_{\ell,i}^\mathcal{T} \leftarrow a_{\ell,i}^\mathcal{T} * 2$ & 84.37 & 82.52 & 87.55 & 84.96 & 21.6 & 6.8 \\
             $\mathrm{Ratio}\leftarrow\sum{a_\mathcal{T}}/\sum{a_\mathcal{V}}$ & 84.20 & 81.57 & 87.56 & 84.46 & 23.1 & 8.2 \\
             \cc \textbf{Ours} & \cc \textbf{84.91} & \cc \textbf{82.84} & \cc \textbf{88.07} & \cc \textbf{85.37} & \cc \textbf{20.0} & \cc \textbf{6.2}  \\
            \bottomrule
        \end{tabular}
    }
    \vspace{-5pt}
    \caption{\textbf{Different Strategies for textual enhancement}. We conduct experiments with different textual enhancement strategies.}
    \label{tab:different strategies}
\end{table}

\section{Conclusion}
In this work, we introduce ONLY, a novel training-free approach that leverages a single additional Transformer layer to mitigate hallucinations in Large Vision-Language Models (LVLMs). By utilizing text-to-visual entropy, ONLY selectively activates attention heads with a high language bias to generate textually-enhanced outputs. These outputs are then adaptively decoded alongside the original output, using either contrastive or collaborative decoding. Extensive evaluations across six benchmarks and three LVLM backbones show that ONLY consistently outperforms existing methods in reducing hallucinations. Moreover, our approach incurs minimal additional inference time and memory consumption, making it well-suited for real-world applications that require real-time responses.

\clearpage

\section*{Acknowledgements}
This work has been funded in part by the Army Research Laboratory (ARL) award W911NF-23-2-0007 and W911QX-24-F-0049, DARPA award FA8750-23-2-1015, and ONR award N00014-23-1-2840. MM is supported by Meta AI Mentorship program and the research is partially supported by National Institutes of Health awards R01MH125740, R01MH132225, and R21MH130767.

{
    \small
    \bibliographystyle{ieeenat_fullname}
    \bibliography{main}
}

\clearpage
\appendix
\setcounter{page}{1}
\maketitlesupplementary

\renewcommand{\thesection}{\Alph{section}}
\renewcommand\thefigure{\Alph{section}\arabic{figure}} 
\renewcommand\thetable{\Alph{section}\arabic{table}}  
\setcounter{section}{0}
\setcounter{figure}{0} 
\setcounter{table}{0} 

This supplementary document is organized as follows:
\begin{itemize}[leftmargin=0.5cm, itemindent=0cm, itemsep=4pt,topsep=4pt,parsep=0pt]
    \item The intuitive and theoretical explanation for our motivation is provided in Section~\ref{sec:motivation of entropy}.
    \item Additional experimental details, including further implementation details, descriptions of other implemented baselines, and license information for the utilized code and datasets, are provided in Section~\ref{sec:detail}.
    \item Additional experimental results on different benchmarks are presented in Section~\ref{sec:more experiments}.
    \item Additional ablation studies with different parameters are presented in Section~\ref{sec:more ablation}.
    \item More case studies and GPT-4V-aided evaluations are provided in Section~\ref{sec:case}.
    \item Potential directions for future work are discussed in Section~\ref{sec:future}.
\end{itemize}

\section{More Explanation on Motivation}
\label{sec:motivation of entropy}
\subsection{Intuitive Explanation for TVER}

Our method is motivated by a principle in information theory: \( H(x|y) \!\leq\! H(x) \). Let \( H(\mathcal{T}) \) and \( H(\mathcal{V}) \) denote the entropy of pure textual and visual attention, respectively. During LVLM decoding, since the model processes both image and text simultaneously, we treat the attention distributions as conditioned on the other modality. This leads to an approximate theoretical form of Eq.~(11): \(\mathrm{TVER} \!=\! \frac{H(\mathcal{T}|\mathcal{V})}{H(\mathcal{V}|\mathcal{T})}\).
Since \( H(\mathcal{T}|\mathcal{V}) \!\leq\! H(\mathcal{T}) \) and \( H(\mathcal{V}|\mathcal{T}) \!\leq\! H(\mathcal{V}) \), a higher \(H(\mathcal{T}|\mathcal{V})\) indicates behavior closer to purely textual inference, while higher \(H(\mathcal{V}|\mathcal{T})\) suggests reliance on visual priors. To approximate the noisy branch used in VCD and M3ID, we aim to enhance textual focus and suppress visual focus, which motivates maximizing \(\mathrm{TVER}\) for effective textual enhancement.

\section{More Experimental Details}
\label{sec:detail}
\subsection{Benchmarks and Metrics}
We conduct extensive experiments on the following benchmarks:
\begin{itemize}
    \item \looseness=-1 \textbf{POPE~\citep{li2023evaluating}} is a popular benchmark for assessing object hallucinations in LVLMs. It tests the models with yes-or-no questions regarding the presence of specific objects, such as, ``\texttt{Is there a \{object\} in the image?}'' The images from the benchmark derive from three existing datasets: MSCOCO~\citep{lin2014microsoft}, A-OKVQA~\citep{schwenk2022okvqa}, and GQA~\citep{hudson2019gqa}, and comprises three distinct subsets—\textit{random}, \textit{popular}, and \textit{adversarial}—based on how the negative samples are generated. For each dataset setting, the benchmark provides 6 questions per image, resulting in 3,000 test instances. We evaluate the performance of different methods using four metrics: accuracy, precision, recall, and F1 score.
    \item \textbf{CHAIR~\citep{rohrbach2018object}} evaluates object hallucinations through image captioning, where the LVLMs are prompted to describe 500 randomly selected images from the MSCOCO validation set. The performance is evaluated based on two metrics:
    \begin{align}
    \label{eq:chair metrics}
        \text{CHAIR}_I &= \frac{\text{\# hallucinated objects}}{\text{\# all objects mentioned}}, \\ \text{CHAIR}_S &=  \frac{\text{\# sentences with hallucinated object}}{\text{\# all sentences}}.
    \end{align}
    \item \textbf{MME-Hallucination~\citep{fu2023mme}} is a comprehensive benchmark consisting of four subsets: \textit{existence} and \textit{count} for object-level hallucinations, and \textit{position} and \textit{color} for attribute-level hallucinations. Each subset includes 30 images and 60 questions, with two questions per image. Similar to POPE~\citep{li2023evaluating}, the benchmark includes yes-or-no questions, and performance is assessed based on binary accuracy. Following the official implementation, the reported score is calculated by combining accuracy and accuracy+, where accuracy is based on individual questions, and accuracy+ is based on images where both questions are answered correctly.
    \item \textbf{MMBench~\citep{liu2025mmbench}} is a comprehensive benchmark designed to evaluate LVLMs' multimodal understanding and reasoning abilities. It emphasizes tasks that require integrating visual and textual information, assessing a model's performance in diverse, real-world scenarios. MMBench employs a hierarchical ability taxonomy, categorizing Perception and Reasoning as Level-1 (L-1) abilities. This taxonomy is further refined into six Level-2 (L-2) dimensions and twenty Level-3 (L-3) dimensions, providing a detailed framework for assessment.
    \item \textbf{MMVP~\citep{tong2024eyes}} is a benchmark designed to assess the fine-grained visual recognition capabilities of LVLMs using CLIP-blind pairs. It comprises 150 image pairs, each paired with a binary-option question. Each image is evaluated separately, and an LVLM's response is deemed correct only if it answers both questions associated with a pair accurately.
    \item \textbf{MM-Vet~\cite{yu2024mmvet}} is a benchmark for evaluating LVLMs on complex tasks. It defines 6 core vision-language capabilities, including recognition, OCR, knowledge, language generation, spatial awareness, and math. An LLM-based evaluator is used to ensure consistent evaluation across diverse question types. The dataset includes 187 images from various online sources and collects 205 questions, each of which requires one or more capabilities to answer.
    \item \textbf{LLaVA-Bench\footnote{\url{https://huggingface.co/datasets/liuhaotian/llava-bench-in-the-wild}.}} includes 24 images depicting complex scenes, memes, paintings, and sketches, accompanied by 60 challenging questions. Selected examples from this dataset are used for qualitative comparisons of responses generated by different decoding methods. Additionally, following \citet{yin2023woodpecker}, we evaluate the accuracy and level of detail in the generated responses using the advanced LVLM, GPT-4V\footnote{\url{https://openai.com/index/gpt-4v-system-card}.}.
\end{itemize}

\subsection{More Implementation Details}
In our experiments, we adhere to the default query format for the input data used in both LLaVA-1.5~\citep{liu2023visual}, InstructBLIP~\citep{dai2024instructblip}, and Qwen-VL~\cite{bai2023qwen}. We set $\alpha_1 = 3$, $\alpha_2 = 1$ by default in our decoding process. Additionally, we set $\gamma = 0.2$ for LLaVA-1.5 and $\gamma = 0.4$ for InstructBLIP/Qwen-VL. We follow VCD~\citep{leng2024mitigating} to implement adaptive plausibility constraints~\citep{li2023contrastive}:
\begin{gather}
p_{\theta}(y_t) = 0, \quad \text{if} \,\,\, y_t \notin \mathcal{S}(y_{<t}),
    \label{eq:adaptive}
\end{gather}
where \(\,\,\,\mathcal{S}(y_{<t}) = \{y_t \in \mathcal{S}: p_{\theta}(y_t | v, \mathbf{x}, \mathbf{y}_{<t}) \geq \beta \max_w  p_{\theta}(w | v, \mathbf{x}, \mathbf{y}_{<t})\}\).
Here, $\mathcal{S}$ is the whole vocabulary of LVLM, and hyperparameter $\beta \in [0, 1]$ controls the truncation of the next token distribution. A larger $\beta$ indicates more aggressive truncation, keeping only the high-probability tokens. In our implementation, we set the logits for $y_t \notin \mathcal{S}(y_{<t})$ to $-\infty$. By default, we set $\beta=0.1$ for all tasks. All experiments are conducted on a single 48GB NVIDIA RTX 6000 Ada GPU. 

\subsection{Pilot Study Details}
For Figure~\ref{fig:feedback}, we visualize 500 images from the CHAIR~\cite{rohrbach2018object} benchmark (left) and 3,000 images from POPE~\cite{li2023evaluating} (right). For Figure~\ref{fig:tver}, we analyze 3,000 POPE images to examine the relationship between entropy deviation and noise level.

\subsection{Devision of Textual and Visual Tokens}
In Eq.~\ref{eq:texual/visual attention}, textual and visual attention are obtained based on the indices corresponding to each modality. The index ranges for both modalities are listed below:
\begin{itemize}
\item LLaVA-1.5~\citep{liu2023visual}: \\Textual indices – [0:35], [611:]; Visual indices – [35:611].
\item InstructBLIP~\cite{dai2024instructblip}: \\Textual indices – [32:]; Visual indices – [0:32].
\item Qwen-VL\cite{bai2023qwen}: \\Textual indices – [257:]; Visual indices – [1:257].
\end{itemize}

\subsection{Details of Other Baselines}
In this work, we mainly compare the performance of our ONLY with two state-of-the-art contrastive-decoding approaches: VCD~\citep{leng2024mitigating} and M3ID~\citep{favero2024multi}. The method and implementation details for these approaches are provided below:
\begin{itemize}
\item \textbf{VCD~\citep{leng2024mitigating}} contrasts output distributions derived from original and distorted visual inputs. Specifically, given a textual query ${x}$ and a visual input ${v}$, the model generates two distinct output distributions: one conditioned on the original ${v}$ and the other conditioned on the distorted visual input ${v'}$, which is obtained by applying pre-defined distortions (e.g., Gaussian noise mask) to ${v}$. 
Then, a new contrastive probability distribution is computed as:
\begin{gather}
p_{vcd}\left(y_t\right) = \text{softmax}[ (1+\alpha) 
f_\theta\left(y | v, \mathbf{x}, \mathbf{y}_{<t}\right) - \nonumber\\ \alpha f_\theta\left(y | v', \mathbf{x}, \mathbf{y}_{<t}\right)].
\end{gather}
In our implementation, we follow the default setting in VCD~\citep{leng2024mitigating} and set $\alpha=1$ for reproduction. To generate $v'$, we use a total of 500 noise steps.

\item \textbf{M3ID~\citep{favero2024multi}} contrasts output distributions derived from original visual inputs with those from pure text inputs, which lack visual information. The final probability distribution is given by:
\begin{gather}
    p_{m3id}\left(y_t\right) = \text{softmax}[f_\theta\left(y | v, \mathbf{x}, \mathbf{y}_{<t}\right) + \nonumber\\ \frac{1 - e^{-\lambda t}}{e^{-\lambda t}} \left( f_\theta\left(y | v, \mathbf{x}, \mathbf{y}_{<t}\right) - f_\theta\left(y | \mathbf{x}, \mathbf{y}_{<t}\right)\right) ].
\end{gather}
Following their recommended best practice, we set the hyperparameter $\lambda$, which balances the conditioned and unconditioned models, to $0.02$.
\end{itemize}

\subsection{Dataset and Code Licensing}
\textbf{Datasets}. We list the known license information for the datasets below: POPE~\citep{li2023evaluating} and MMVP~\citep{tong2024eyes} benchmarks are licensed under MIT License.
CHAIR~\citep{rohrbach2018object} is made available under the BSD 2-Clause License.
LLaVA-Bench is available under Apache-2.0 License.
MME-Hallucination~\citep{fu2023mme} benchmark dataset is collected by Xiamen University for academic research only. MM-Vet~\cite{yu2024mmvet} dataset is under the CC BY-NC 4.0 license.

\textbf{Code}. In this work, we also use some code implementations from the existing codebases: LLaVA~\citep{liu2023visual} and VCD~\citep{leng2024mitigating} are licensed under the Apache-2.0 License.
InstructBLIP~\citep{dai2024instructblip} is under BSD-3-Clause License.
Qwen-VL~\cite{bai2023qwen} is under the Tongyi Qianwen License.

\section{More Experimental Results and Analysis}
\label{sec:more experiments}

\begin{table*}[t!]
        \begin{center}
        \begin{small}
        \setlength{\tabcolsep}{6pt} 
        \caption{\looseness=-1\RebuttalRevision{\textbf{Results on MME-Hallucination~\citep{fu2023mme} benchmark.} We report the average MME scores along with the standard deviation across three random seeds for each subset. We also report the total scores achieved by the different methods across all four subsets in the final column. Higher scores ($\uparrow$) indicate better performance. The best results  are \textbf{bolded}, and the second-best are \underline{underlined}.}}
        \label{tab:MME-full}
        \vspace{-5pt}
        \resizebox{\textwidth}{!}{
        \begin{tabular}{llccccc}
            \toprule
             \multirow{2}{*}[-0.5ex]{\textbf{Model}} & \multirow{2}{*}[-0.5ex]{\textbf{Method}}  &  \multicolumn{2}{c}{\textbf{Object-level}} & \multicolumn{2}{c}{\textbf{Attribute-level}} & \multirow{2}{*}[-0.5ex]{\textbf{Total Score $\uparrow$}}\\
             \cmidrule(lr){3-4}\cmidrule(lr){5-6}
              &   & Existence $\uparrow$ & Count $\uparrow$ & Position $\uparrow$ & Color $\uparrow$   & \\
             \midrule
            \multirow{8}{*}{{\textbf{LLaVA-1.5}}} & Regular &  173.75 {\tiny ($\pm4.79$)} & 121.67 {\tiny ($\pm12.47$)} & 117.92 {\tiny ($\pm3.69$)\phantom{0}} & 149.17 {\tiny ($\pm7.51$)\phantom{0}} &  562.50 {\tiny ($\pm3.96$)\phantom{0}} \\
             & DoLa  & 176.67 {\tiny ($\pm2.89$)} & 113.33 {\tiny ($\pm10.41$)} & 90.55 {\tiny ($\pm8.22$)} & 141.67 {\tiny ($\pm7.64$)\phantom{0}} &  522.22 {\tiny ($\pm16.78$)}  \\
             & OPERA  & 183.33 {\tiny ($\pm6.45$)} & \underline{137.22} {\tiny ($\pm6.31$)\phantom{0}} & 122.78 {\tiny ($\pm2.55$)\phantom{0}} & \underline{155.00} {\tiny ($\pm5.00$)\phantom{0}}  &  \underline{598.33} {\tiny ($\pm10.41$)}  \\
             & VCD  & {186.67} {\tiny ($\pm5.77$)} & 125.56 {\tiny ($\pm3.47$)\phantom{0}} & 128.89 {\tiny ($\pm6.73$)\phantom{0}} & 139.45 {\tiny ($\pm12.51$)} &  580.56 {\tiny ($\pm15.13$)}  \\
             & M3ID  & {186.67} {\tiny ($\pm5.77$)} &  128.33 {\tiny ($\pm10.41$)} & \underline{131.67} {\tiny ($\pm5.00$)\phantom{0}} & 151.67 {\tiny ($\pm20.88$)} &  598.11 {\tiny ($\pm20.35$)}  \\
             & {Woodpecker} & {\underline{187.50} {\tiny ($\pm2.89$)}} & {125.00 {\tiny ($\pm0.00$)\phantom{0}}} & {126.66 {\tiny ($\pm2.89$)\phantom{0}}} & {149.17 {\tiny ($\pm17.34$)}} &  {588.33 {\tiny ($\pm10.00$)}} \\
            & {HALC} & {183.33 {\tiny ($\pm0.00$)}} & {133.33 {\tiny ($\pm5.77$)\phantom{0}}} & {107.92 {\tiny ($\pm3.69$)\phantom{0}}} &{\underline{155.00} {\tiny ($\pm5.00$)\phantom{0}}}  &  {579.58 {\tiny ($\pm9.07$)\phantom{0}}} \\
             & \cc \textbf{Ours} & \cc \textbf{191.67} {\tiny ($\pm2.89$)} & \cc \textbf{145.55} {\tiny ($\pm10.72$)\phantom{0}}  & \cc \textbf{136.66} {\tiny ($\pm2.89$)\phantom{0}} & \cc \textbf{161.66} {\tiny ($\pm2.89$)\phantom{0}} & \cc \textbf{635.55} {\tiny ($\pm5.85$)\phantom{0}} \\
            \midrule
            \multirow{6}{*}{{\textbf{InstructBLIP}}} & Regular & 160.42 {\tiny ($\pm5.16$)} & 79.17 {\tiny ($\pm8.22$)\phantom{0}} & \textbf{79.58} {\tiny ($\pm8.54$)\phantom{0}} & \underline{130.42} {\tiny ($\pm17.34$)}  &  \underline{449.58} {\tiny ($\pm24.09$)}    \\
             & DoLa  & 175.00 {\tiny ($\pm5.00$)} & 55.00 {\tiny ($\pm5.00$)\phantom{0}} & 48.89 {\tiny ($\pm3.47$)\phantom{0}} & 113.33 {\tiny ($\pm6.67$)\phantom{0}}  &  392.22 {\tiny ($\pm7.88$)\phantom{0}}   \\
             & OPERA  & 175.00 {\tiny ($\pm3.33$)} & 61.11 {\tiny ($\pm3.47$)\phantom{0}} & 53.89 {\tiny ($\pm1.92$)\phantom{0}} &120.55 {\tiny ($\pm2.55$)\phantom{0}}  &  410.56 {\tiny ($\pm9.07$)\phantom{0}}     \\
             & VCD  & 158.89 {\tiny ($\pm5.85$)} &  \textbf{91.67} {\tiny ($\pm18.34$)} & 66.11 {\tiny ($\pm9.76$)\phantom{0}} & 121.67 {\tiny ($\pm12.58$)} & 438.33 {\tiny ($\pm16.07$)}  \\
             & M3ID  & 160.00 {\tiny ($\pm5.00$)} & \underline{87.22} {\tiny ($\pm22.63$)} & 69.44 {\tiny ($\pm9.18$)\phantom{0}} & 125.00 {\tiny ($\pm7.64$)\phantom{0}} & 441.67 {\tiny ($\pm17.32$)}    \\
             & \cc \textbf{Ours} & \cc \textbf{180.00} {\tiny ($\pm5.00$)} & \cc {77.78} {\tiny ($\pm7.70$)\phantom{0}}  & \cc \underline{74.44} {\tiny ($\pm12.05$)\phantom{0}} & \cc \textbf{135.55} {\tiny ($\pm3.85$)\phantom{0}} & \cc \textbf{467.77} {\tiny ($\pm8.55$)} \\
             \midrule
            \multirow{4}{*}{{\textbf{Qwen-VL}}} & Regular & 155.00 {\tiny ($\pm3.54$)} & 127.67 {\tiny ($\pm13.36$)} & {131.67} {\tiny ($\pm7.73$)\phantom{0}} & 173.00 {\tiny ($\pm9.75$)\phantom{0}}  &  587.33 {\tiny ($\pm31.06$)}    \\
             & VCD  & 156.00 {\tiny ($\pm6.52$)} &  131.00 {\tiny ($\pm6.19$)\phantom{0}} & 128.00 {\tiny ($\pm3.61$)\phantom{0}} & \textbf{181.67} {\tiny ($\pm5.14$)\phantom{0}} & 596.67 {\tiny ($\pm11.61$)}  \\
             & M3ID  & \underline{178.33} {\tiny ($\pm2.89$)} & \underline{143.33} {\tiny ($\pm2.89$)\phantom{0}} & \underline{150.00} {\tiny ($\pm2.89$)\phantom{0}} & 175.00 {\tiny ($\pm5.00$)\phantom{0}} & \underline{646.66} {\tiny ($\pm8.50$)\phantom{0}}    \\
             & \cc \textbf{Ours} & \cc \textbf{180.00} {\tiny ($\pm5.00$)} & \cc \textbf{146.67} {\tiny ($\pm5.00$)\phantom{0}}  & \cc \textbf{156.11} {\tiny ($\pm6.31$)\phantom{0}} & \cc \underline{178.33} {\tiny ($\pm2.89$)\phantom{0}} & \cc \textbf{661.11} {\tiny ($\pm3.47$)\phantom{0}} \\
            \bottomrule
        \end{tabular}
        }
        \vspace{-15pt}
        \end{small}
        \end{center}
        
\end{table*}

\subsection{Full Results on MME-Hallucination}
In Table~\ref{tab:MME-full}, we present the full results on the MME-Hallucination benchmark. From the results, our method consistently outperforms others on both object-level and attribute-level data across three LVLM backbones.

\begin{table}[t]
\centering
\resizebox{\linewidth}{!}{
\begin{tabular}{lccccccc}
\toprule
Method & LR & AR & RR & FP-S & FP-C & CP & Overall \\
\midrule
Regular & 30.51 & 71.36 & 52.17 & 67.58 & \textbf{58.74} & 76.35 & 64.09 \\
VCD     & 30.51 & \textbf{73.37} & 53.04 & \textbf{67.92} & 57.34 & 77.03 & 64.60 \\
M3ID    & 30.51 & 72.36 & 53.04 & 67.58 & 57.34 & \textbf{77.36} & 64.43 \\
\cc \textbf{Ours}    & \cc \textbf{33.05} & \cc \textbf{73.37} & \cc \textbf{54.78} & \cc {66.55} & \cc \textbf{58.74} & \cc \textbf{77.36} & \cc \textbf{64.95} \\
\bottomrule
\end{tabular}
}
\vspace{-5pt}
\caption{
\textbf{Detailed results on MMBench benchmark}. Abbreviations adopted: LR for Logical Reasoning; AR for Attribute Reasoning; RR for Relation Reasoning; FP-S for Fine-grained Perception (Single Instance); FP-C for Fine-grained Perception (Cross Instance); CP for Coarse Perception. The best results are \textbf{bolded}.}
\vspace{5pt}
\label{tab:MMBench Full}
\end{table}

\subsection{Full Results on MMBench}
In Table~\ref{tab:MMBench Full}, we present the overall performance on the MMBench benchmark, as well as the detailed performance across six Level-2 abilities: Logical Reasoning (LR), Attribute Reasoning (AR), Relation Reasoning (RR), Fine-grained Perception - Single Instance (FP-S), Fine-grained Perception - Cross Instance (FP-C), and Coarse Perception (CP). We follow VCD~\citep{leng2024mitigating} to conduct experiments on the MMBench-\texttt{dev} set. Our method outperforms other baselines in most abilities and the overall score.

\begin{table}[t]
\centering
\resizebox{\linewidth}{!}{
\begin{tabular}{lccccccc}
\toprule
Method & Rec & OCR & Know & Gen & Spat & Math & Total \\
\midrule
Regular & 30.8 & 19.0 & 14.5 & 17.9 & 26.9 & \textbf{11.5} & 26.1 \\
VCD     & 35.6 & {21.9} & 18.3 & \underline{21.9} & {28.9} & 3.8 & {30.9} \\
M3ID    & 35.0 & 19.7 & {18.8} & 19.0 & 26.0 & {7.7} & 29.9 \\
DoLA  & \underline{37.2} & 22.1 & 17.9 & 21.0 & 26.3 & 7.7 & 31.7 \\
OPERA  & 35.4 & \textbf{25.6} & \underline{20.5} & \textbf{22.9} & \underline{30.9} & 11.5 & \underline{32.0} \\
HALC  & 36.2 & 21.5 & 17.5 & 20.1 & 23.5 & \textbf{7.7} & 30.8 \\
\cc \textbf{Ours} & \cc \textbf{37.3} & \cc \underline{23.9} & \cc \textbf{22.9} & \cc \underline{22.1} & \cc \textbf{31.3} & \cc {3.8} & \cc \textbf{32.8} \\
\bottomrule
\end{tabular}
}
\vspace{-5pt}
\caption{
\textbf{Detailed results on MM-Vet benchmark}. Abbreviations adopted: Rec for Recognition, OCR for Optical Character Recognition, Know for Knowledge, Gen for Language Generation, Spat for Spatial Awareness, Math for Mathematics. The best results are \textbf{bolded}, and the second best are \underline{underlined}.}
\label{tab:mmvet}
\end{table}

\subsection{Results on MM-Vet}
In Table~\ref{tab:mmvet}, we present the overall performance on the MM-Vet~\citep{yu2024mmvet} benchmark, where we use LLaVA-1.5 as the LVLM backbone. From the results, we observed that our method consistently outperforms others on the MM-Vet benchmark.

\subsection{Evaluation on other advanced LVLMs}

We further report results of LLaVA-NeXT-7B \!/\! 13B~\cite{liu2024llavanext} on POPE (MS-COCO) benchmark in table \ref{tab:llavanext}. Our method consistently outperforms existing approaches at both scales while requiring only half the inference time and resources.
\begin{table}[h]
\vspace{-3mm}
\centering
    \resizebox{\linewidth}{!}{
    \begin{tabular}{lcccccccc}
        \toprule
        \multirow{2}{*}[-0.5ex]{\textbf{Method}} & \multicolumn{4}{c}{\textbf{LLaVA-NeXT-7B}} & \multicolumn{4}{c}{\textbf{LLaVA-NeXT-13B}} \\
        \cmidrule(lr){2-5} \cmidrule(lr){6-9}
         & Acc. & Prec. & Rec. & F1 & Acc. & Prec. & Rec. & F1 \\
        \midrule
        Regular & 85.71 & 85.27 & 86.33 & 85.80 & 86.74 & 86.53 & 87.04 & 86.78 \\
        VCD     & 87.07 & 87.40 & 86.62 & 87.01 & 87.09 & {87.39} & 86.69 & 87.04 \\
        M3ID    & {87.48} & {87.64} & \textbf{87.27} & {87.45} & {87.84} & \textbf{87.95} & {87.71} & {87.83} \\
        \cc Ours & \cc \textbf{87.96} & \cc \textbf{88.59} & \cc {87.13} & \cc \textbf{87.86} & \cc \textbf{87.94} & \cc 87.31 & \cc \textbf{88.80} & \cc \textbf{88.05} \\
        \bottomrule
    \end{tabular}}
\vspace{-5pt}
\caption{
\textbf{Detailed results with LLaVA-NeXT}. The best results are \textbf{bolded}, and the second best are \underline{underlined}.}
\label{tab:llavanext}
\end{table}

\section{More Ablation Studies and Analysis}
\label{sec:more ablation}

\subsection{Effects of \texorpdfstring{$\alpha_1$ and $\alpha_2$}{alpha1 and alpha2} in Adaptive Decoding}
In Section~\ref{sec:method}, we introduce collaborative and contrastive decoding, along with hyperparameters $\alpha_1$ and $\alpha_2$, which regulate the influence of the textual-enhanced branch. Tables~\ref{tab:alpha1} and \ref{tab:alpha2} analyze their impact, showing that the default values $\alpha_1=3$ and $\alpha_2=1$ yield the best performance across benchmarks. Notably, setting these to 0 reduces our approach to standard decoding, confirming that adaptive decoding significantly enhances hallucination mitigation in LVLMs.

\begin{table}[t]
\centering
    \resizebox{\linewidth}{!}{
    \begin{tabular}{lcccccccc}
            \toprule
              \multirow{2}{*}[-0.5ex]{\textbf{Values}}  &  \multicolumn{4}{c}{\textbf{POPE}} & \multicolumn{2}{c}{\textbf{CHAIR}} \\
             \cmidrule(lr){2-5}\cmidrule(lr){6-7}
                 & {Acc.}  & {Prec.}  & {Rec.}  & {F1}  & CHAIR$_S$  & CHAIR$_I$ \\
             \midrule
             $\alpha_1 =0$ & 88.13 & \textbf{94.55} & 80.93 & 87.21 & 23.5 & 8.6 \\
             $\alpha_1 =1$ & 88.27 & 94.50 & 81.27 & 87.38 & 22.4 & 7.8 \\
             $\alpha_1 =2$ & 88.87 & 89.63 & 88.10 & 88.86 & 21.5 & 7.2 \\
             \cc $\alpha_1 =3$ & \cc \textbf{89.70} & \cc 89.95 & \cc \textbf{88.27} & \cc \textbf{89.10} & \cc \textbf{20.0} & \cc \textbf{6.2} \\
             $\alpha_1 =4$ & 88.37 & 88.85 & 87.94 & 88.39 & 22.3 & 7.6 \\
            \bottomrule
        \end{tabular}
    }
    \vspace{-5pt}
    \caption{\textbf{Sensitivity analysis of hyperparameter $\alpha_1$}. We present the performance of our approach, based on the LLaVA-1.5 backbone, across two benchmarks for varying values of $\alpha_1$. Note that we fix $\alpha_2=1$ in this experiment.}
    \label{tab:alpha1}
\end{table}

\begin{table}[t]
\centering
%
    \resizebox{\linewidth}{!}{
    \begin{tabular}{lcccccccc}
            \toprule
              \multirow{2}{*}[-0.5ex]{\textbf{Values}}  &  \multicolumn{4}{c}{\textbf{POPE}} & \multicolumn{2}{c}{\textbf{CHAIR}} \\
             \cmidrule(lr){2-5}\cmidrule(lr){6-7}
                 & {Acc.}  & {Prec.}  & {Rec.}  & {F1}  & CHAIR$_S$  & CHAIR$_I$ \\
             \midrule
             $\alpha_2 =0$ & 86.50 & 86.35 & 88.13 & 86.72 & 24.8 & 9.3 \\
             \cc $\alpha_2 =1$ & \cc \textbf{89.70} & \cc 89.95 & \cc \textbf{88.27} & \cc \textbf{89.10} & \cc \textbf{20.0} & \cc \textbf{6.2} \\
             $\alpha_2 =2$ & 87.67 & 96.69 & 78.00 & 86.35 & 22.4 & 7.6 \\
             $\alpha_2 =3$ & 87.37 & \textbf{97.14} & 77.00 & 85.91 & 23.4 & 7.3 \\
             $\alpha_2 =4$ & 87.13 & 97.12 & 76.53 & 85.61 & 24.2 & 8.1 \\
            \bottomrule
        \end{tabular}
    }
    \vspace{-5pt}
\caption{\textbf{Sensitivity analysis of hyperparameter $\alpha_2$}. We present the performance of our approach, based on the LLaVA-1.5 backbone, across two benchmarks for varying values of $\alpha_1$. Note that we fix $\alpha_1=3$ in this experiment.}
    \label{tab:alpha2}
\end{table}

\begin{table}[t]
\centering
%
    \resizebox{\linewidth}{!}{
    \begin{tabular}{lcccccccc}
            \toprule
              \multirow{2}{*}[-0.5ex]{\textbf{Values}}  &  \multicolumn{4}{c}{\textbf{POPE}} & \multicolumn{2}{c}{\textbf{CHAIR}} \\
             \cmidrule(lr){2-5}\cmidrule(lr){6-7}
                 & {Acc.}  & {Prec.}  & {Rec.}  & {F1}  & CHAIR$_S$  & CHAIR$_I$ \\
             \midrule
             $\beta =0$ & 87.70 & 93.40 & 81.13 & 86.84 & 24.6 & 10.1 \\
             $\beta =0.05$ & 88.17 & \textbf{94.21} & 81.33 & 87.30 & 23.7 & 9.6 \\
             \cc $\beta =0.1$ & \cc \textbf{89.70} & \cc 89.95 & \cc \textbf{88.27} & \cc \textbf{89.10} & \cc \textbf{20.0} & \cc \textbf{6.2} \\
             $\beta =0.25$ & 89.56 & 89.48 & 87.63 & 88.55 & 21.4 & 7.6 \\
             $\beta =0.5$ & 89.47 & 89.83 & 86.53 & 88.15 & 22.1 & 7.2 \\
            \bottomrule
        \end{tabular}
    }
    \vspace{-5pt}
\caption{\textbf{Sensitivity analysis of hyperparameter $\beta$}. We present the performance of our approach, based on the LLaVA-1.5 backbone, across two benchmarks for varying values of $\beta$.}
    \label{tab:beta}
\end{table}

\begin{table}[t]
\centering
%
    \resizebox{\linewidth}{!}{
    \begin{tabular}{lcccccccc}
            \toprule
              \multirow{2}{*}[-0.5ex]{\textbf{Values}}  &  \multicolumn{4}{c}{\textbf{POPE}} & \multicolumn{2}{c}{\textbf{CHAIR}} \\
             \cmidrule(lr){2-5}\cmidrule(lr){6-7}
                 & {Acc.}  & {Prec.}  & {Rec.}  & {F1}  & CHAIR$_S$  & CHAIR$_I$ \\
             \midrule
             $\gamma =0.0$ & 89.13 & 90.41 & 86.38 & 88.35 & 23.5 & 8.2 \\
             $\gamma =0.1$ & 89.20 & 89.88 & 86.73 & 88.28 & 22.6 & 8.1 \\
             \cc $\gamma =0.2$ & \cc \textbf{89.70} & \cc 89.95 & \cc \textbf{88.27} & \cc \textbf{89.10} & \cc \textbf{20.0} & \cc \textbf{6.2} \\
             $\gamma =0.3$ & 89.40 & 93.20 & 85.00 & 88.91 & 21.2 & 7.1 \\
             $\gamma =0.4$ & 89.03 & 93.99 & 83.40 & 88.38 & 21.7 & 7.0 \\
             $\gamma =0.5$ & 89.15 & 92.26 & 84.29 & 88.10 & 22.4 & 7.6 \\
             $\gamma =0.6$ & 89.21 & 91.78 & 85.39 & 88.47 & 23.1 & 8.1 \\
            \bottomrule
        \end{tabular}
    }
    \vspace{-5pt}
    \caption{\textbf{Sensitivity analysis of hyperparameter $\gamma$}. We present the performance of our approach, based on the LLaVA-1.5 backbone, across two benchmarks for varying values of $\gamma$.}
    \label{tab:gamma}
\end{table}

\subsection{Effect of \texorpdfstring{$\beta$}{beta} in Adaptive Plausibility Constraint}
We perform an ablation study on $\beta$, introduced in Eq.~\ref{eq:adaptive}, by varying its value from 0 to 0.5 while keeping all other hyperparameters fixed. As shown in Table~\ref{tab:beta}, setting $\beta=0$, which removes the constraint, leads to suboptimal performance across both benchmarks. Our method achieves the best results with $\beta=0.1$, which we adopt as the default setting.

\subsection{Effect of \texorpdfstring{$\gamma$}{gamma} in Adaptive Plausibility Constraint}
We further studied the influence led by the threshold $\gamma$ for adaptive decoding. The results in Table~\ref{tab:gamma} show that setting $\gamma=0.2$ reaches the optimal result for LLaVA-1.5. Besides, we keep $\gamma=0.4$ for other baseline LVLMs.

\subsection{Scaling Up the LVLMs}
We extend our evaluation to the 13B variant of the LLaVA-1.5 model to assess the scalability of our approach. Table~\ref{tab:POPE13} compares our results with state-of-the-art methods across all three subsets of the POPE benchmark using the 13B model. Our findings show that increasing model size does not mitigate hallucination issues, as the 7B and 13B models exhibit comparable performance. Notably, ONLY consistently outperforms other approaches across all subsets, demonstrating its effectiveness and scalability.

\subsection{Details about Ablation Studies on Layer Selection and Strategies}
In Section \ref{sec:ablation on layer}, we conduct two ablation studies to validate our proposed method. Detailed results are provided below.

\textbf{Selection of Layer for Textual Enhancement:} In this experiment, we select a single layer from the total of 32 layers for textual enhancement. The F1 scores for our method across the 32 layers are as follows:
[85.37, 85.20, 84.74, 84.7, 85.11, 84.68, 85.17, 84.69, 85.18, 84.73, 84.7, 84.74, 84.9, 84.94, 84.88, 85.06, 84.62, 84.67, 84.83, 85.15, 84.72, 84.76, 84.99, 85.03, 84.7, 84.93, 84.76, 84.9, 84.8, 85.12, 84.62, 84.76].  
In comparison, the results for regular decoding, VCD~\cite{leng2024mitigating}, and M3ID~\cite{favero2024multi} are 81.27, 83.38, and 84.05, respectively.

\textbf{Other Strategies for Textual Enhancement:} In Table~\ref{tab:different strategies}, we explore additional strategies for textual enhancement, which include:

\begin{itemize}
    \item $a_{\ell,i}^\mathcal{V} \leftarrow 0$: Setting the visual attention in the attention matrix to zero, inspired by M3ID~\cite{favero2024multi}, which uses a visual-free input for contrastive decoding;
    \item $a_{\ell,i}^\mathcal{V} \leftarrow a_{\ell,i}^\mathcal{V} + \varepsilon$: Adding noise $\varepsilon$ to the visual attention, inspired by VCD~\cite{leng2024mitigating}, which uses a distorted visual input for contrastive decoding;
    \item $a_{\ell,i}^\mathcal{T} \leftarrow a_{\ell,i}^\mathcal{T} * 2$: Enhancing textual attention by directly multiplying it by 2;
    \item $\mathrm{Ratio}\leftarrow\sum{a_\mathcal{T}}/\sum{a_\mathcal{V}}$: Instead of using the text-to-visual entropy ratio as the criterion to select textual-enhanced heads, we use the ratio between the sum of textual attention and visual attention. Heads with a ratio lower than the average across all heads are masked out, as described in Eq.~\ref{eq:select head}.
\end{itemize}

All of these strategies require minimal additional computation, providing an efficiency advantage over other methods~\cite{leng2024mitigating, favero2024multi}. This demonstrates the effectiveness of using just one layer for mitigating hallucinations in LVLMs, rather than relying on an extra full-process inference.

\begin{table}[t]
    \centering
    \small
    \caption{
        \textbf{Results on POPE~\citep{li2023evaluating} benchmark using 13B-sized LLaVA-1.5}. Higher ($\uparrow$) accuracy, precision, recall, and F1 indicate better performance. 
    }
    \vspace{5pt}
    \label{tab:POPE13}
    \resizebox{\linewidth}{!}{
    \begin{tabular}{cclcccc}
    \toprule
     & \multirow{2}{*}[-2pt]{\textbf{Setup}} & \multirow{2}{*}[-2pt]{\textbf{Method}} & \multicolumn{4}{c}{\textbf{LLaVA-1.5}}  \\
    \arrayrulecolor{gray} \cmidrule(lr){4-7} 
     &  &  & {Acc.} $\uparrow$ & {Prec.} $\uparrow$ & {Rec.} $\uparrow$ & {F1} $\uparrow$  \\
    \midrule
    \multirow{15}{*}[-5pt]{\rotatebox{90}{\textbf{\normalsize MS-COCO}}} & \multirow{5}{*}{Random} 
    & Regular &  82.53 & 78.57 & 89.47 & 83.67 \\
     &  & VCD  &  84.80 & 80.67 & 91.53 & 85.76 \\
     &  & M3ID  & 85.37  & 81.30 & 91.87 & 86.26  \\
     &  & \cc \textbf{Ours} &\cc \textbf{88.63} &\cc \textbf{89.66} &\cc 87.33 &\cc \textbf{88.48}  \\
     \arrayrulecolor{gray}\cmidrule(lr){2-7}
      &  \multirow{5}{*}{Popular} & Regular & 80.53 & 76.17 & 88.87 & 82.03\\
     &  & VCD  &  82.23 & 76.88 & 92.20 & 83.84 \\
     &  & M3ID  &  82.60 & 77.91 & 91.00 & 83.95 \\
     &  & \cc \textbf{Ours} &\cc \textbf{85.47} &\cc \textbf{83.25}  &\cc 88.80 &\cc \textbf{85.94}  \\
     \arrayrulecolor{gray}\cmidrule(lr){2-7}
      &  \multirow{5}{*}{Adversarial} & Regular & 75.80 & 70.41 & 89.00 & 78.62 \\
     &  & VCD  &  77.33 & 71.44 & 91.07 & 80.07 \\
     &  & M3ID  & 77.43  & 71.65 & 90.80 & 80.09 \\
     &  & \cc \textbf{Ours} &\cc \textbf{80.63}   &\cc \textbf{76.33} &\cc 88.80 &\cc \textbf{82.10}    \\
    \bottomrule
    \end{tabular}
}
\end{table}

\section{More Case Studies}
\label{sec:case}
\subsection{Details about GPT-4V-Aided Evaluation}
Following VCD~\citep{leng2024mitigating}, we use GPT-4V to evaluate responses in open-ended generation scenarios, scoring them based on accuracy and detailedness. Leveraging GPT-4V's strong human-like capabilities, it can detect incorrect colors, positions, and relationships, allowing for a thorough evaluation of the responses.

Specifically, we apply the prompt in Table~\ref{tab:prompt_evaluation} to instruct GPT-4V to rate two responses on a scale from 1 to 10 for both accuracy and detailedness:
\begin{itemize}
    \item \textbf{Accuracy} measures the consistency between the responses/descriptions generated by the LVLMs and the given image. A lower score is given if GPT-4V detects any inconsistencies in the content.
    \item \textbf{Detailedness} evaluates the depth and specificity of the responses. A higher score is awarded if the response includes comprehensive descriptions, captures fine-grained details of the image, and provides well-elaborated explanations. Conversely, a lower score is given if the response is vague or lacks sufficient detail.
\end{itemize}

\begin{table*}[t]\centering
\begin{minipage}{0.95\textwidth}
\centering
\begin{tcolorbox} 
    \centering
   
      \small
    \begin{tabular}{p{0.95\textwidth}} \hline \\
   \textbf{Description:} \\    
   
   AI that scores image description accuracy and detailedness.

   \\ \midrule

   \textbf{Instructions:} \\   
   
You are an AI designed to evaluate and score the performance of two AI assistants in describing a given image. Your primary focus is on the accuracy and detailedness of their descriptions. You will assess the accuracy by checking for hallucinations - any part of the description that is inconsistent with the image content. For detailedness, you will consider how rich the response is in necessary details, excluding any hallucinated parts. You will provide scores on a scale from 1 to 10 for each assistant separately, based on these criteria. After scoring, you will offer an explanation for your evaluation, ensuring it is free from bias and not influenced by the order of presentation of the responses.
\\ \\
Input format: \\ \\
\lbrack{}Assistant 1\rbrack{}\\
 \{Response 1\}  \\
\lbrack{}End of Assistant 1\rbrack{} \\
\\
\lbrack{}Assistant 2\rbrack{} \\
 \{Response 2\}\\
\lbrack{}End of Assistant 2\rbrack{} \\
\\
Output format:\\
\\
Accuracy:\\
Scores of the two answers:\\
Reason:\\
\\
Detailedness:\\
Scores of the two answers:\\
Reason:\\ \\

\bottomrule
    \end{tabular}
\end{tcolorbox}
\caption{\textbf{GPT-4V-aided evaluation setup}. We present the prompt we provided to GPT-4V to evaluate the LVLM responses based on accuracy and detailedness.}
\label{tab:prompt_evaluation}
\end{minipage}
\end{table*}

\subsection{More Qualitative Results}
In Figure~\ref{fig:llavabench2}, we present additional case studies from LLaVA-Bench to qualitatively demonstrate the effectiveness of our methods in mitigating hallucinations. We also include GPT-4V evaluations, providing accuracy and detailedness scores for each instance.

\begin{figure}[h]

\includegraphics[width=\linewidth]{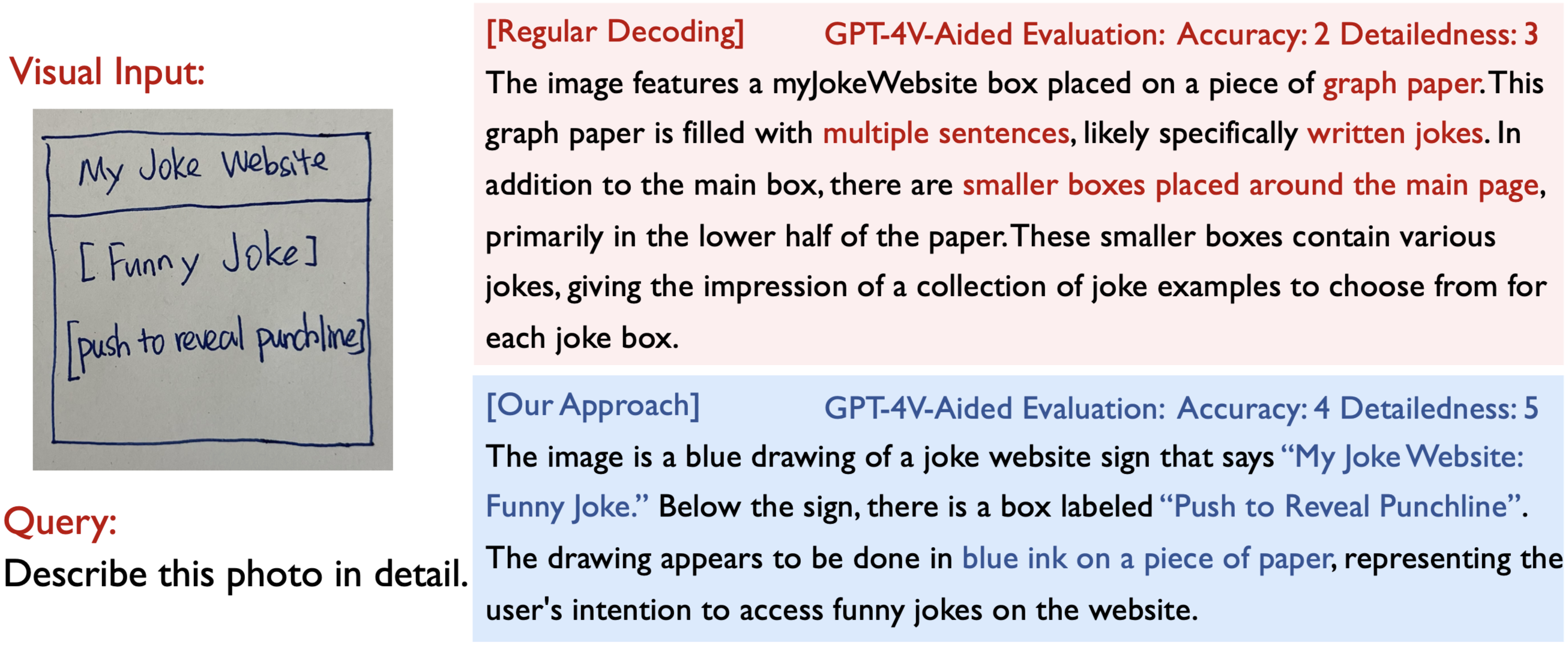}
\\[10pt]
\includegraphics[width=\linewidth]{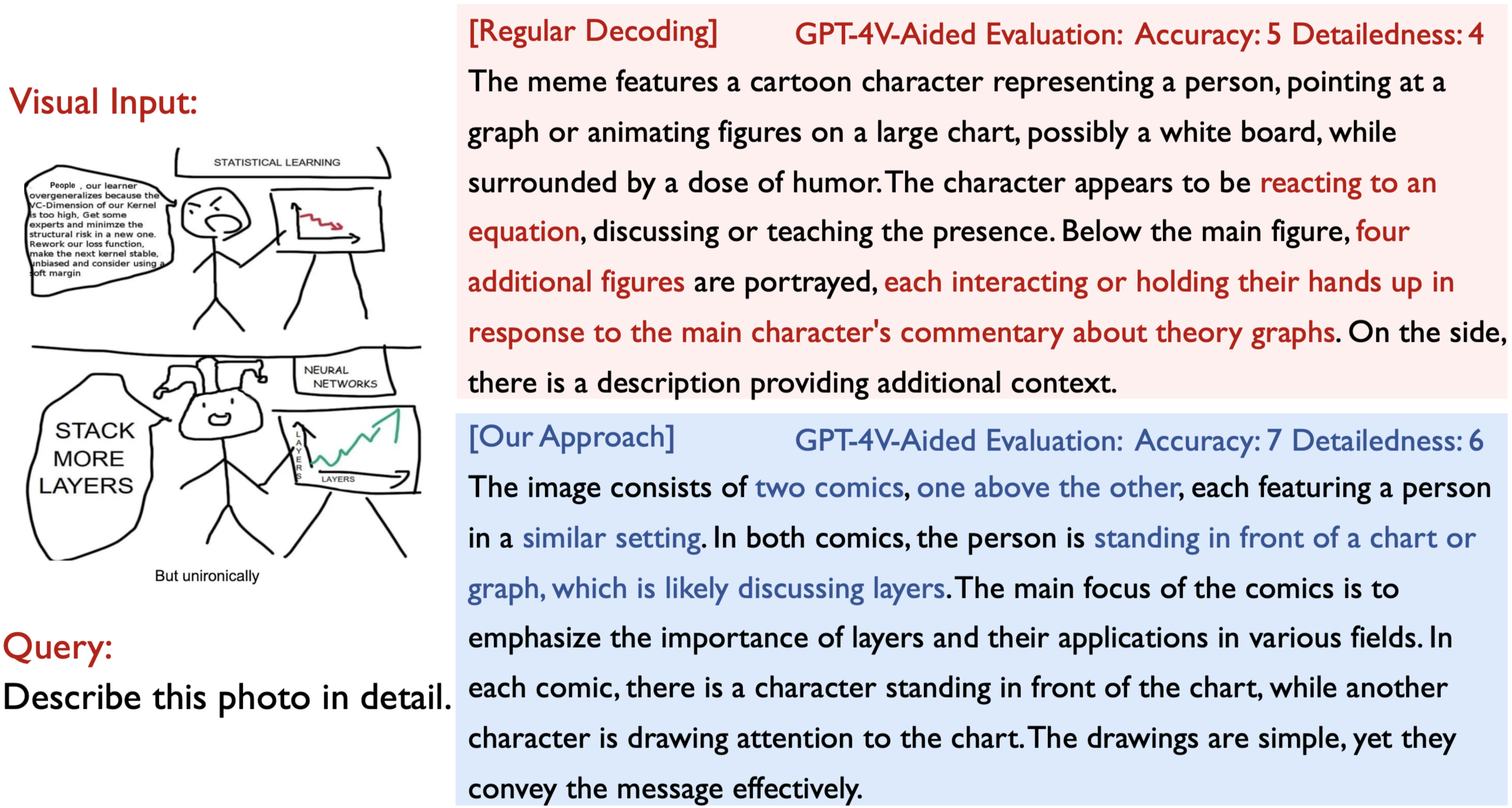}
\\[10pt]
\includegraphics[width=\linewidth]{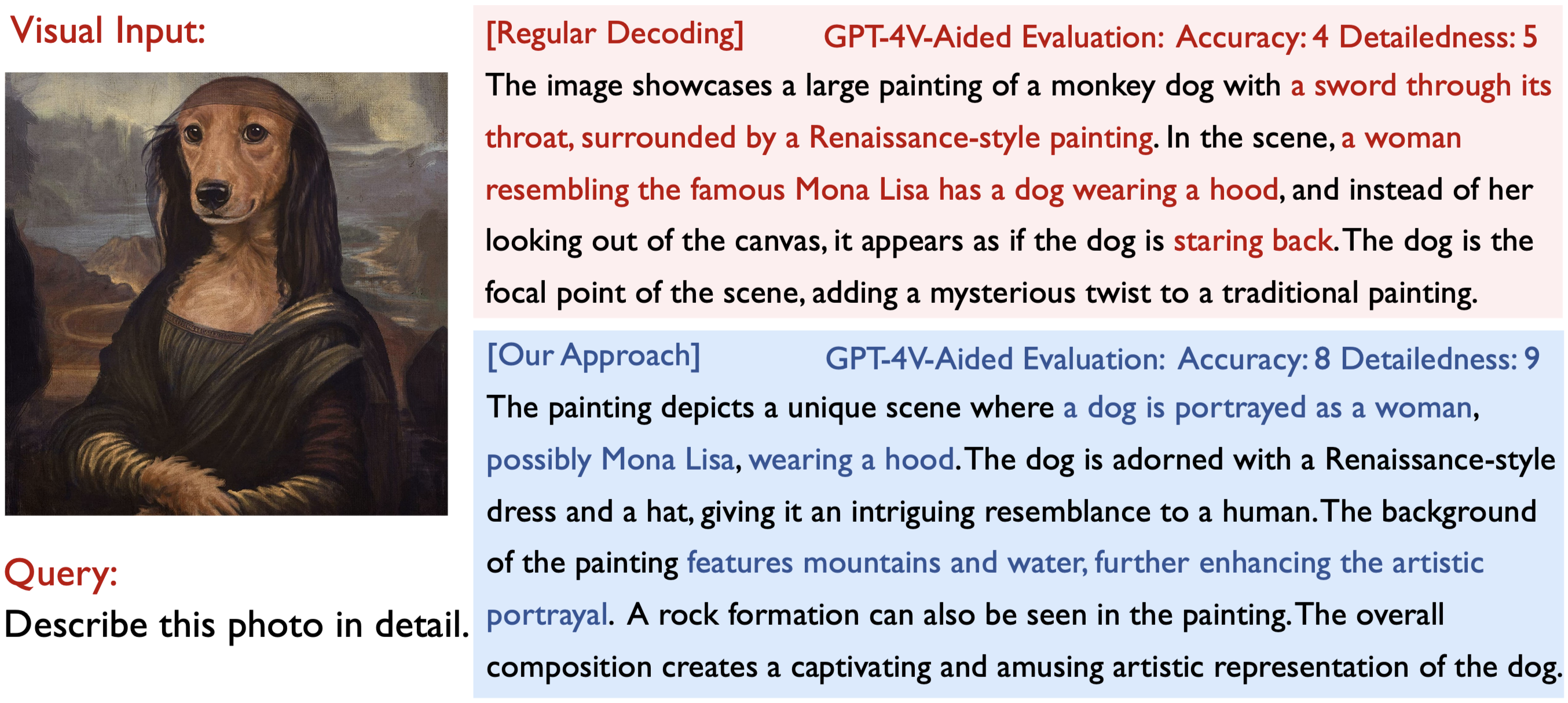}

\caption{\textbf{Case studies on the LLaVA-Bench benchmark}. We compare the responses generated by regular decoding and our method using LLaVA-1.5. GPT-4V-aided evaluation results are also provided alongside the responses. Hallucinated and accurate content is highlighted in \textcolor{darkred}{red} and \textcolor{darkblue}{blue}.} 

\label{fig:llavabench2}
\end{figure}

\section{Future Work}
\label{sec:future}
In future work, we aim to further improve the speed of our method and develop a more efficient hallucination mitigation approach that surpasses the original LVLM speed, leveraging efficient LVLM techniques like FastV~\cite{chen2024image} and VScan~\cite{zhang2025vscan}. Additionally, we plan to explore our method's potential for video hallucination mitigation to demonstrate its adaptability across various tasks.

\end{document}